\documentclass[10pt,journal,compsoc]{IEEEtran}
\usepackage{amsmath,amsfonts}
\usepackage{algorithmic}
\usepackage{algorithm}
\usepackage{array}
\usepackage[caption=false,font=normalsize,labelfont=sf,textfont=sf]{subfig}
\usepackage{textcomp}
\usepackage{stfloats}
\usepackage{url}
\usepackage{verbatim}
\usepackage{graphicx}
\usepackage{cite}
\usepackage[pagebackref=true,breaklinks=true,letterpaper=true,colorlinks,citecolor=blue,linkcolor=blue,bookmarks=false]{hyperref}

\begin{document}

\title{GPHM: Gaussian Parametric Head Model for Monocular Head Avatar Reconstruction}

\author{
Yuelang Xu, 
Zhaoqi Su,
Qingyao Wu,
Yebin Liu
\thanks{
\indent Yuelang Xu, Zhaoqi Su, and Yebin Liu are with the Department of Automation, Tsinghua University, Beijing 100084, P.R.China.\\
\indent Qingyao Wu is with the School of Software Engineering, South China University of Technology, Beijing, P.R.China.\\
\indent Corresponding author: Yebin Liu.}
}

\newcommand{\mname}{ACLRNet}
\markboth{subimit to IEEE Transactions on Pattern Analysis and Machine Intelligence,~Vol.~XX, No.~XX, XX~2024}%
{Shell \MakeLowercase{\textit{Zamir et al.}}: Bare Demo of IEEEtran.cls for Computer Society Journals}

\IEEEtitleabstractindextext{%
\begin{abstract}
Creating high-fidelity 3D human head avatars is crucial for applications in VR/AR, digital human, and film production. Recent advances have leveraged morphable face models to generate animated head avatars from easily accessible data, representing varying identities and expressions within a low-dimensional parametric space. However, existing methods often struggle with modeling complex appearance details, e.g., hairstyles, and suffer from low rendering quality and efficiency. In this paper we introduce a novel approach, 3D Gaussian Parametric Head Model, which employs 3D Gaussians to accurately represent the complexities of the human head, allowing precise control over both identity and expression. The Gaussian model can handle intricate details, enabling realistic representations of varying appearances and complex expressions. 
Furthermore, we presents a well-designed training framework to ensure smooth convergence, providing a robust guarantee for learning the rich content. 
Our method achieves high-quality, photo-realistic rendering with real-time efficiency, making it a valuable contribution to the field of parametric head models.
Finally, we apply the 3D Gaussian Parametric Head Model to monocular video or few-shot head avatar reconstruction tasks, which enables instant reconstruction of high-quality 3D head avatars even when input data is extremely limited, surpassing previous methods in terms of reconstruction quality and training speed.
Project page: \url{https://yuelangx.github.io/gphmv2/}.
\end{abstract}

\begin{IEEEkeywords}
Gaussian Splatting, Parametric Model, Head Avatar
\end{IEEEkeywords}
}
\IEEEdisplaynontitleabstractindextext

\maketitle

\section{Introduction}
\label{sec:intro}

\IEEEPARstart{C}{reating} high-fidelity 3D human head avatars holds significant importance across various fields, including VR/AR, telepresence, digital human interfaces, and film production. The automatic generation of such avatars has been a focal point in computer vision research for many years. Recent methods~\cite{zhao2023havatar, xu2023latentavatar, xu2023avatarmav, gao2022reconstructing, zielonka2022instant, zheng2022imavatar, zheng2023pointavatar, grassal2022neural, gafni2021dynamic, qin2023high} can create an animated head avatar through conveniently collected data such as a monocular video data or even a picture~\cite{li2023goha, khakhulin2022rome}. Serving as the most fundamental tool in these methods, the 3D morphable models (3DMM)~\cite{gerig2018morphable, li2017learning}, which represent varying identities and expressions within a low-dimensional space, have been proven to be a highly successful avenue in addressing this challenging problem.

Since the traditional parametric 3DMMs are typically limited by the topology of the underlying template mesh and only focus on the face part, some works~\cite{yenamandra2020i3dmm, lin2023ssif, giebenhain2023nphm, giebenhain2023mononphm} propose to use implicit Signed Distance Field (SDF) as the geometric representation to model the entire head. Despite their flexibility, these methods fall short in recovering high-frequency geometric and/or texture details like complex hairstyles, glasses or accessories. 
On the other end of the spectrum, Neural Radiance Field (NeRF)~\cite{mildenhall2020nerf} based methods~\cite{zhuang2022mofanerf, hong2022headnerf} learn parametric head models by directly synthesizing photo-realistic images, thus eliminating the need of geometry modeling. However, NeRF is built upon volumetric rendering, which involves sampling and integrating points distributed throughout space. Therefore, NeRF-based methods typically suffer from low rendering efficiency and have to trade it off with rendering resolution, thereby greatly reducing rendering quality. Moreover, skipping geometric reconstruction would probably lead to poor 3D consistency.



More recently, 3D Gaussian Splatting (3DGS)~\cite{kerbl3Dgaussians}, which uses explicit Gaussian ellipsoids to represent 3D scenes, has attracted significant attention from the research community. Experiments have verified the superior quality of the rendered results and excellent rendering efficiency compared to previous NeRF-based or surface-based methods even on dynamic scenes~\cite{wu20234d, yang2023realtime, luiten2023dynamic, yang2023deformable}. 
Motivated by this progress, we propose a novel \textbf{3D Gaussian Parametric Head Model} for head avatar modeling, which, for the first time, marries the power of 3DGS with the challenging task of parametric head modeling. Our 3D gaussian parametric head model decouples the control signals of the head into the latent spaces of identity and expression, as is also done in SDF-based face model NPHM~\cite{giebenhain2023nphm}. 
These latent spaces are then mapped to the offsets of the Gaussian positions, which effectively represent the variance of shape and appearance of different identities and expressions. Benefiting from the differentiability of Gaussian splatting, our model can be learned from multi-view video data corpus in an end-to-end manner, without relying on geometry supervision, achieving high quality monocular head avatar reconstruction results. 

Unfortunately, training our 3D Gaussian parametric head model is not quite straightforward, because Gaussian ellipsoids are unstructured and each Gaussian ellipsoid has its own independent learnable attribute. Such a characteristic makes 3DGS powerful in overfitting a specific object or scene, but poses great challenges for generative head modeling. Without proper initialization and regularization, the learned parametric head model may suffer from unstable training or a large number of Gaussian points becoming redundant and noisy, as shown in Fig.~\ref{fig:ablation_initializaiton}. 

To overcome these challenges,
we propose a well-designed two-stage training strategy to ensure smooth convergence of our model training. Specifically, we first roughly train all the networks on a mesh-based guiding model. Subsequently, the network parameters are migrated to the Gaussian model, and all Gaussian points are initialized with the trained mesh geometry to ensure that they are located near the actual surface. Compared to naive initialization with  FLAME~\cite{li2017learning}, our initialization strategy leads to a better guess of the positions of Gaussian points, making the subsequent training of the model converge stably and the areas like hairs better recovered. Moreover, we propose to use 3D landmark loss to supervise the deformation of the model learning expressions, which can speed up the convergence and avoid artifacts under exaggerated expressions. 
Lastly, our method supports training from both 3D head scans and multi-view 2D face datasets, which enhances the versatility and comprehensiveness of facial data collection and model training.

After training on large corpus of multi-view head videos, our parametric Gaussian head model can generate photorealistic images that accurately depict the diverse range of facial appearances, naturally handling complex and exaggerated expressions, while also enabling real-time rendering. Additionally, our method supports single-image fitting and surpasses previous techniques in both reconstruction accuracy and identity consistency. Furthermore, the model resulting from our fitting process allows for the control of various expressions while maintaining naturalness and consistent identity even under exaggerated expressions.

A preliminary version of this work has been published in ECCV 2024~\cite{xu2024gphm}, in which we propose a novel 3D Gaussian Parametric Head Model (GPHM) enabling photo-realistic representation of human heads and high-quality face avatar from a single image. However, the preliminary work~\cite{xu2024gphm} mainly focuses on representing a parametric head model, lacking ease of use and robustness for downstream tasks like head reconstruction from input images. In the current version, we present GPHMv2, a head avatar reconstruction framework, which supports instant and robust head avatar reconstruction from monocular video or even few-shot image inputs. The proposed head reconstruction pipeline surpasses previous NeRF-based~\cite{xu2023avatarmav, gao2022reconstructing, zielonka2022instant, xu2023latentavatar} or 3DGS-based~\cite{shao2024splattingavatar, xiang2024flashavatar} head reconstruction methods in reconstruction quality and training speed. Moreover, previous methods heavily rely on the 3DMM models~\cite{li2017learning, gerig2018morphable}, suffering from the coupling of expression and shape, and perform poorly in cross-identity reenactment (see Sec~\ref{sec::experiments}). And due to lacking sufficient prior information, these methods can hardly support few-shot or one-shot head reconstruction like our method.

We extend the preliminary version~\cite{xu2024gphm} as follows. Firstly, we extend our network structure and adjust data preprocessing for a more expressive and generalizable head model. Specifically, to better disentangle the expression and head motion of the avatar and capture more detailed expression information, we introduce a facial expression encoder and a non-face motion encoder to extract latent expressions and latent motions from images (see Section~\ref{subsec:model_representation},~\ref{subsec:loss_functions}). These components are jointly trained with the other network components to enable end-to-end avatar animation via image reconstruction results. 
However, training the model in an end-to-end self-reconstruction manner and directly using the input image as the expression condition inevitably leads to the leakage of identity-related appearance information into the latent expression codes. To address this, we utilize LivePortrait~\cite{guo2024liveportrait} to synthesize a large number of images with different identities but the same expression as the additional expression condition during training, effectively eliminating this issue (see Section~\ref{subsec:preprocessing}). Therefore, the model can more accurately isolate and capture expression features independently of identity, leading to more precise and versatile avatar animations.

Secondly, we enhance the functionality of the preliminary model by designing a few-shot 3D head avatar reconstruction framework. Leveraging the pre-trained extended GPHMv2 model, a high-quality 3D head avatar can be rapidly reconstructed using only a small amount of monocular data. Specifically, we optimize our model for a single identity-specific avatar in a two-stage process, from coarse to fine. In the first stage, we focus solely on optimizing the identity code to provide a rough initialization. In the second stage, we refine the model by optimizing the neutral Gaussian attributes and motion-related networks to capture finer details. Also in this stage, we introduce a tiny network to accurately model the expression-related dynamic changes in color and Gaussian attributes. Finally, the well-finetuned head avatar can be driven by any video fed into the encoders.

The contributions of our method can be summarized as:
\begin{itemize}

\item We propose 3D Gaussian Parametric Head Model, a novel parametric head model which utilizes 3D Gaussians as the representation, enabling photo-realistic rendering quality and real-time rendering speed.


\item We propose a well-designed training strategy to ensure that the Gaussian model converges stably while learning rich appearance details and complex expressions efficiently.


\item We extend our preliminary proposed 3D Gaussian Parametric Head Model to the more expressive and generalizable GPHMv2, which enables instant head avatar reconstruction from monocular video or even few-shot images, achieving state-of-the-art quality and training time.

\end{itemize}


\begin{figure*}[t]
  \centering
  \includegraphics[width=\linewidth]{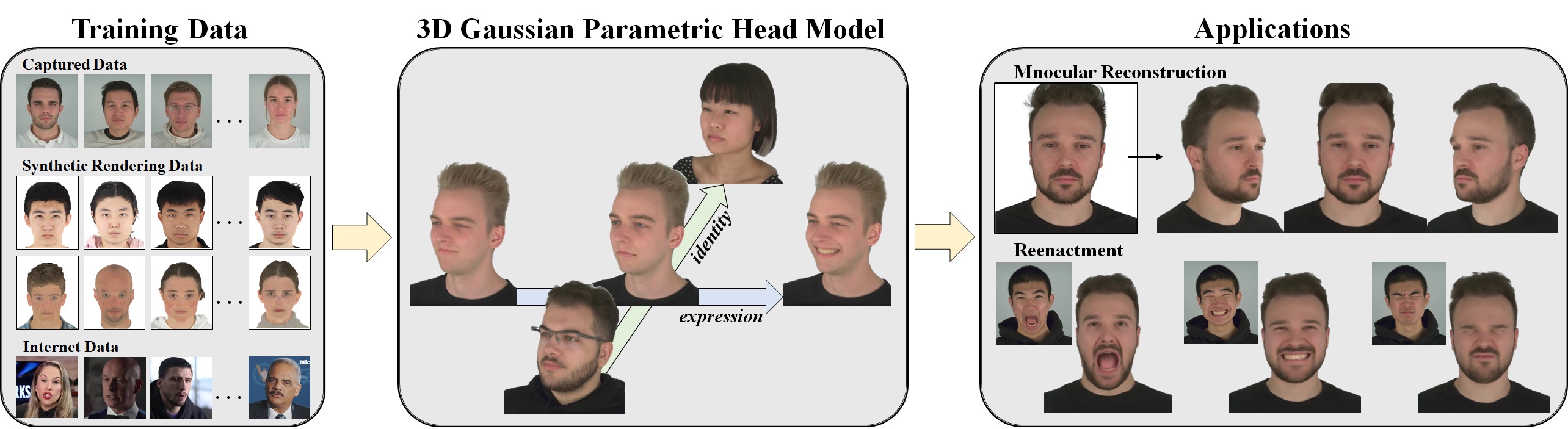}
  \caption{We utilize hybrid datasets comprising captured multi-view video data and rendered image data from 3D scans for training our model. The trained model can be manipulated using decoupled identity and expression codes to produce a diverse array of high-fidelity head models. When presented with an image, our model can be adjusted to reconstruct the portrait in the image and edit the expression according to any other desired expressions.}
  \label{fig:teaser}
\end{figure*}

\section{Related Work}
\label{sec:related}

\noindent\textbf{Parametric Head Models.}
Parametric head models are used to represent facial features, expressions, and identities effectively and efficiently. They allow for the creation of realistic human faces with adjustable parameters, making them essential in computer graphics, animation, and virtual reality. Therefore, research in this field has always been a hot topic. Traditional 3D Morphable Models(3DMM)~\cite{blanz1999morphable, li2017learning, gerig2018morphable, cao2014facewarehouse, wang2022faceverse} are constructed by non-rigidly registering a template mesh with fixed topology to a series of 3D scans. Through this registration process, a 3DMM can be computed using dimensionality reduction techniques such as principal component analysis (PCA). The resulting parametric space captures the variations in facial geometry and appearance across a population. However, while 3DMMs offer a powerful way to represent faces, they do have limitations. These models rely heavily on the correspondence between the 3D scans and the template for accurate fitting and may struggle to represent local surface details like wrinkles or hair styles that deviate significantly from the template mesh. 
Recent advances in implicit representation have led to the great development of neural parametric head models. Some methods~\cite{yenamandra2020i3dmm, wu2023ganhead, giebenhain2023nphm, giebenhain2023mononphm} propose implicit Signed Distance Field (SDF) based head models, which are not constrained by topology thus can recover more complex content like hair compared to previous mesh-based Methods. Other methods~\cite{zhuang2022mofanerf, hong2022headnerf, wang2022morf, buhler2023preface} propose to use NeRF~\cite{mildenhall2020nerf} as the representation of the parametric head models, which can directly synthesize photorealistic images without geometric reconstruction. Cao, et al.~\cite{cao2022authentic} use a hybrid representation~\cite{lombardi2021mixture} of mesh and NeRF to train their model on unpublished large-scale light stage data. However, rendering efficiency is typically low in NeRF-based methods, often resulting in a trade-off with rendering resolution.

\noindent\textbf{3D GAN based Head Models.}
3D Generative Adversarial Networks (GANs) have revolutionized the field of computer vision, particularly in the domain of human head and face modeling, enabling the generation of face avatars from input images. Traditional methods often require labor-intensive manual work or rely on multi-view images to create 3D models. 3D GANs as a more automated and data-driven approach, which are just trained on single-view 2D images but generate detailed and realistic 3D models of human head~\cite{chan2022efficient, Chan2020pigan, gu2022stylenerf, Roy2021stylesdf, deng2021gram, xiang2022gramhd}. Panohead~\cite{an2023panohead} additionally introduces images of hairstyles on the back of characters and trains a full-head generative model. Based on the previous methods, IDE-3D~\cite{sun2022ide} proposes to use semantic map to edit the 3D head model. Next3D~\cite{sun2023next3d} and AniFaceGAN~\cite{yue2022anifacegan} extend to uses the FLAME model~\cite{li2017learning} to condition the generated head model, so that the expression and pose of the generated head model can be controlled. AniPortraitGAN~\cite{yue2023aniportraitgan} further replaces FLAME model with SMPLX model~\cite{pavlakos2019expressive} to generate upper body avatars, thus the shoulders and the neck can also be controlled.
These 3D GAN-based models primarily leverage the coarse FLAME model for expression control, often leading to a loss of expression details in the generated faces. In contrast, our method directly learns the expression distribution from the dataset, capturing more facial appearance details.

\noindent\textbf{3D Gaussians-based Head Models.}
Recently, 3D Gaussian splatting~\cite{kerbl3Dgaussians} has shown superior performance compared to NeRF, excelling in both novel view synthesis quality and rendering speed. Several methods have expanded Gaussian representation to dynamic scene reconstruction~\cite{wu20234d, yang2023realtime, luiten2023dynamic, yang2023deformable}. For human body avatar modeling, recent approaches~\cite{li2023animatable, hu2023gaussianavatar} propose training a 3D Gaussian avatar animated by SMPL~\cite{loper2015smpl} or a skeleton from multi-view videos, surpassing previous methods in rendering quality and efficiency. In the realm of human head avatar modeling, recent techniques~\cite{xu2024gaussian, qian2023gaussianavatars, saito2023relightable, wang2024gaussianhead} also utilize 3D Gaussians to create high-fidelity and efficient head avatars. These approaches centers on the creation of a high-fidelity person-specific avatar using data of a single person. In contrast, our method focus on a versatile prior model that can accommodate varying appearances. Once trained, our model is also capable of person-specific avatar reconstruction by fitting to the input image data provided.

\noindent\textbf{Monocular 3D Head Avatar Reconstruction.}
3D head avatars reconstruction from monocular videos is also a popular yet challenging research topic. Early methods~\cite{cao2015real, cao2016real, ichim2015dynamic, hu2017avatar, deng2019accurate, nagano2018pagan} optimize a morphable mesh to fit the training video. Recent methods~\cite{grassal2022neural, Khakhulin2022realistic} leverage neural networks to learn non-rigid deformation upon 3DMM face templates~\cite{li2017learning, gerig2018morphable}, thus can recover more dynamic details. However, such methods are not flexible enough to handle complex topologies. IMavatar~\cite{zheng2022imavatar} proposes to learn head avatars with implicit SDF-based geometry~\cite{park2019deepsdf, occupancy2019mescheder}, thus getting rid of the topology limitation of the mesh templates. PointAvatar~\cite{zheng2023pointavatar} combines the explicit point cloud with the implicit representation to improve the quality of the rendered images. As NeRF~\cite{mildenhall2020nerf} demonstrates its ability to synthesize high-fidelity novel view images, several methods~\cite{guo2021ad, liu2022semantic, gafni2021dynamic, athar2021flame, athar2022rignerf, xu2023latentavatar, qin2023high} attempt to exploit such representation for neural head modeling. Furthermore, the voxel-based data structure is introduced for training acceleration~\cite{gao2022reconstructing, xu2023avatarmav, zielonka2022instant}. However, these representations usually suffer from the loss of high-frequency details. To overcome, recent methods~\cite{shao2024splattingavatar, xiang2024flashavatar, chen2023monogaussianavatar} introduce 3D Gaussian representation~\cite{kerbl3Dgaussians} to model the head avatars, thereby improving the reconstruction quality while reducing the training time requirement. However, despite these advancements, there is still potential for further enhancement in reconstruction quality and training/inference speed.
\section{Method}
\label{sec:method}

\begin{figure*}[t]
  \centering
  \includegraphics[width=\linewidth]{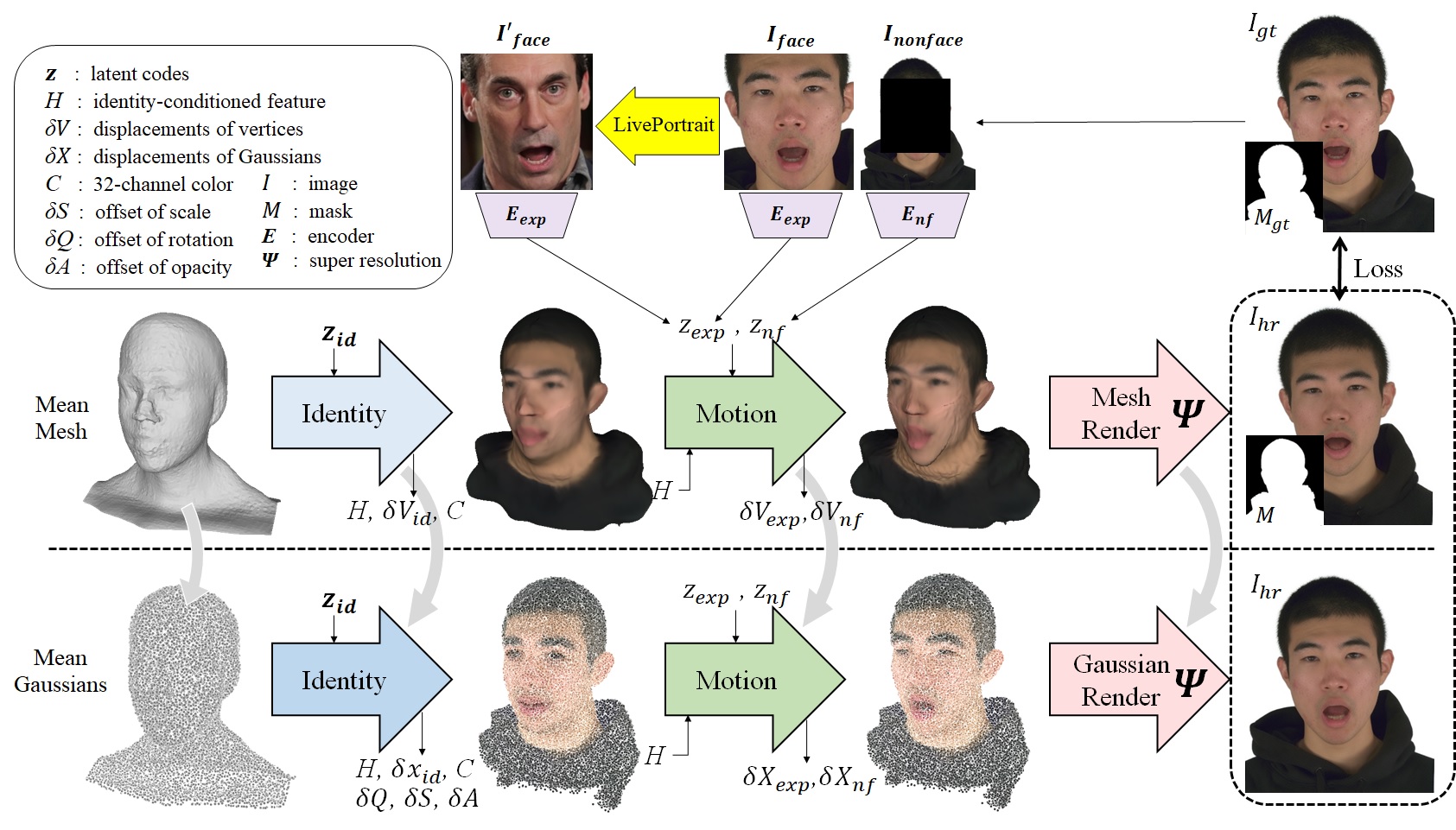}
  \caption{The overview of our GPHM model. Our training strategy can be divided into a Guiding Geometry Model for initialization, and a final 3D Gaussian Parametric Head Model. Deformations of each model are further decoupled into identity-related, expression-related and non-face deformations. For the expression condition images, we input crop groundtruth face image or synthesized images via LivePortrait~\cite{guo2024liveportrait}. For the non-face motion condition, we input groundtruth images with the face area masked. The renderer involves a convolutional refine network $\boldsymbol{\Psi}$, which finally transfers the feature maps from mesh/Gaussian renderer to fine portrait images. During inference, our output exclusively comes from the Gaussian model.}
  \label{fig:method}
\end{figure*}


In this section, we present the 3D Gaussian Parametric Head Model. In contrast to previous mesh-based or NeRF-based models, initializing and training Gaussian-based models pose distinct challenges. This section introduces the dataset and preprocessing, the carefully designed guiding geometry model, the Gaussian Parametric Head Model, and outlines their respective training processes. Additionally, we will provide the training details and demonstrate how to utilize our model for head avatar reconstruction when given an input monocular video.

\begin{figure}[t]
  \centering
  \includegraphics[width=\linewidth]{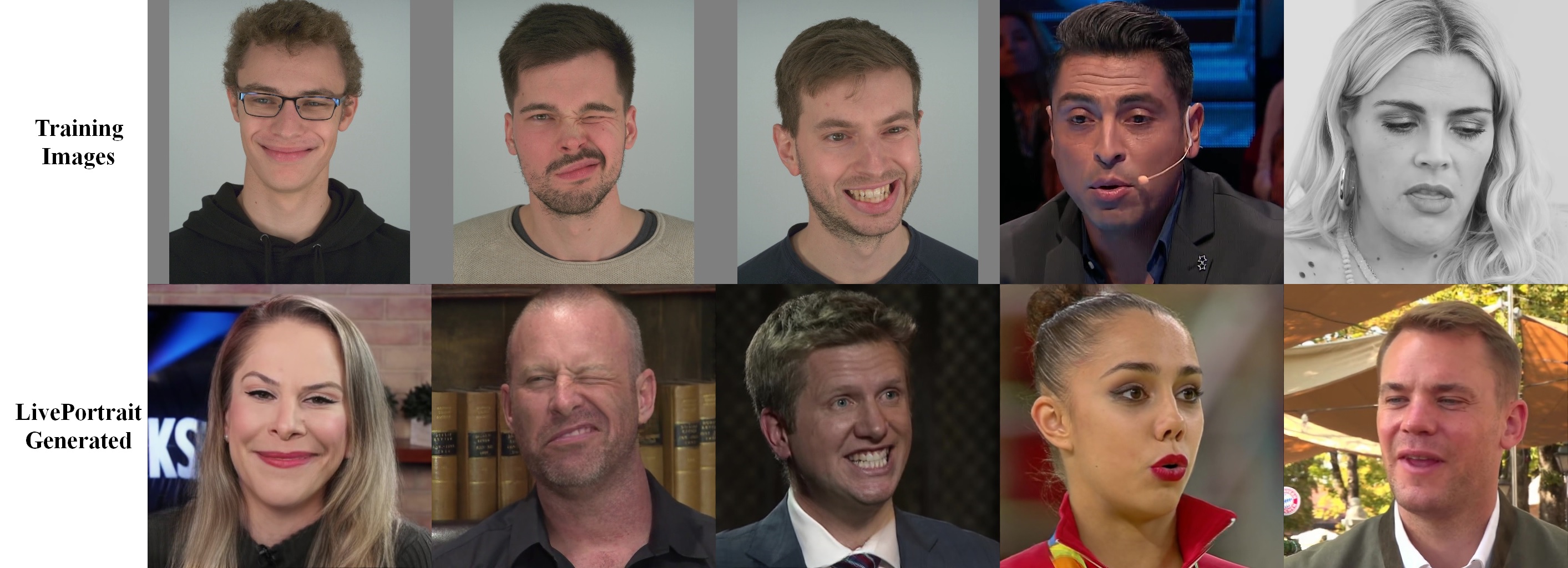}
  \caption{We generate additional expression condition images via LivePortrait~\cite{guo2024liveportrait} for training the appearance decoupled expression encoder.}
  \label{fig:liveportrait}
\end{figure}

\subsection{Data Preprocessing}
\label{subsec:preprocessing}
We used 4 datasets for our model training, including a multi-view video dataset NeRSemble~\cite{kirschstein2023nersemble}, a large-scale monocular video dataset VFHQ~\cite{xie2022vfhq}, two 3D scans datasets NPHM~\cite{giebenhain2023nphm} and FaceVerse~\cite{wang2022faceverse}. We do not use the 3D geometry of the scans directly, but render them into multi-view images and use only the images from the 4 datasets as supervision. To better utilize these 4 different datasets, preprocessing is necessary. First, we resize the images to 512 resolution and adjust the camera parameters. Note that for the monocular videos in VFHQ dataset, we assign a global default value to the camera parameters.

Then, we use BackgroundMattingV2~\cite{lin2021real} to extract the foreground characters in the NeRSemble dataset and record the masks. For the VFHQ dataset, we use RobustVideoMatting~\cite{shanchuan2021robust} to segment foreground and masks. This step is not required for the two synthetic datasets. Next, we use face alignment~\cite{bulat2017how} to detect 2D landmarks in all the images. Through these 2D landmarks, we fit a Basel Face Model (BFM)~\cite{gerig2018morphable} for each expression of each identity, and record the head pose and 3D landmarks of the BFM. For all the images in which the facial area is visible, we extract the face region images through the 2D landmarks and the non-face images by masking the face region. 

Finally, we use LivePortrait~\cite{guo2024liveportrait} to synthesize additional expression condition images. Specifically, for each frontal face image in the training set, we randomly sample another from VFHQ. The original image provides the expression as the driving image, and the sampled image provides the appearance as the source image, creating a new image with the same expression but a random identity. All synthesized images are used as additional expression condition images to train the expression encoder for better generalization as described in Section~\ref{subsec:model_representation}.

We will use the processed camera parameters, images, masks, head pose, 3D landmarks, face images, non-face images, and synthesized expression condition images mentioned above to train our parametric head model.

\subsection{Model Representation}
\label{subsec:model_representation}

The representation of 3D Gaussians poses challenges due to its unordered and unstructured nature, leading to difficulties in the continuous spread of gradients to neighboring points in space during backpropagation. This often results in convergence failure when Gaussians are randomly initialized. On the other hand, surface-based representations such as mesh are just suitable for rough geometry learning. A direct idea is to utilize an existing 3DMM, such as FLAME~\cite{li2017learning}, as the initial position for the points in 3D Gaussian splatting~\cite{kerbl3Dgaussians}. However, this coarse initialization still fails to converge the positions of 3D points to the correct locations, as shown in Fig.~\ref{fig:ablation_initializaiton}. The network tends to alter the shape of the ellipsoid to achieve a suitable fitting result, leading to inaccurate geometry of the point cloud and blurriness in the rendered image.

To address this problem, a more detailed initialization process is necessary for capturing the diverse head variations using 3D Gaussian splatting. Specifically, we draw inspiration from Gaussian Head Avatar~\cite{xu2024gaussian} and leverage the implicit signed distance field (SDF) representation to train a guiding geometry model. This guiding geometry model serves as the initial value for the Gaussian model, providing a more effective starting point for the optimization process. We define the initial model as Guiding Geometry Model and the refined model as 3D Gaussian Parametric Head Model. 

In addition, setting a separate expression code for each frame of data to model dynamic motion like the preliminary version~\cite{xu2024gphm} will cause facial expressions to be coupled with body parts. And as more data is added, it becomes more difficult to optimize a large number of discrete latent codes.

Therefore, we first use two separate latent codes: facial expression codes and non-face motion codes to control facial expressions and non-facial movements respectively. Furthermore, we introduce two additional motion encoders accordingly to avoid directly optimizing those latent codes. A face encoder extracts facial expression codes from face images, while a non-face encoder extracts non-face motion codes from non-face images. Moreover, the 2 encoders can be directly used to extract motion from driving images via a single forward pass in subsequent tasks.

\textbf{Guiding Geometry Model.} The guiding geometry model receives an identity code $\boldsymbol{z^{id}}$, a face image $I_{face}$ for expression condition and a non-face image $I_{nonface}$ for controlling non-face area as input, producing a mesh with vertices $V$, faces $F$, and per-vertex color $C$ that aligns with the specified identity and expression. To achieve this, we use an MLP denoted as $\boldsymbol{f_{mean}}(\cdot)$ to implicitly model the SDF, which represents the mean geometry: 
\begin{equation}
 s, \gamma = \boldsymbol{f_{mean}}(x),
\end{equation}
where $s$ denotes the SDF value, $\gamma$ denotes the feature from the last layer and $x$ denotes the input position. Then, we convert the implicit SDF through Deep Marching Tetrahedra (DMTet)~\cite{shen2021dmtet} into an explicit mesh with vertices positions $V_{0}$, per-vertex feature $\Gamma$ and faces $F$. 
Next, we need to transform the mean shape into a neutral-expression shape on condition of the input identity code $\boldsymbol{z^{id}}$. To inject identity information into the vertices of the mesh, we first use an injection MLP $\boldsymbol{f_{inj}}(\cdot)$, which takes the identity code $\boldsymbol{z^{id}}$ and the per-vertex feature $\Gamma$ as input and produces the identity-conditioned per-vertex feature vectors $H = \boldsymbol{f_{inj}}(\boldsymbol{z^{id}}, \Gamma)$.
Subsequently, utilizing a tiny MLP $\boldsymbol{f_{id}}(\cdot)$, we predict the displacement $\delta V_{id}$ for each vertex. This displacement is used to transform the mean shape into the neutral-expression shape conditioned on the id code $\boldsymbol{z^{id}}$.
\begin{equation}
 \delta V_{id} = \boldsymbol{f_{id}}(H).
\end{equation}

After completing deformations related to identity, the next step is to capture the deformation induced by facial expressions and head pose. Here, the same as GHA~\cite{xu2024gaussian}, we define the human face area as the canonical reference system. In addition to facial expression changes, we need to consider the movement of non-face areas such as the neck and body relative to the head while the face is rigidly transformed. Specifically, we introduce 2 motion encoders $\boldsymbol{E}_{exp}(\cdot)$ for face area, $\boldsymbol{E}_{nf}(\cdot)$ for non-face area, and 2 tiny MLPs $\boldsymbol{f_{exp}}(\cdot)$ for face area and $\boldsymbol{f_{nf}}(\cdot)$ for non-face area. The facial expression encoder takes the condition face image $I_{face}$ as input and predict the facial expression code $z^{exp}$. The non-face motion encoder takes the condition non-face image $I_{nonface}$ as input and predict the non-face motion code $z^{nf}$. Note, during training, the face image can be from any one view in the current frame or the synthesized expression condition images. Then, $\boldsymbol{f_{exp}}(\cdot)$ takes the feature vectors $H$ obtained in the previous step and the expression code $z^{exp}$ from the expression encoder as input, and outputs the displacement $\delta V_{exp} = \boldsymbol{f_{exp}}(H, z^{exp})$ for each vertex. $\boldsymbol{f_{nf}}(\cdot)$ takes the feature vectors $H$ and a non-face motion code $z^{nf}$ as input, and outputs the displacement $\delta V_{nf} = \boldsymbol{f_{nf}}(H, z^{nf})$ for each vertex. Using this displacement, we update the vertex positions to $V_{can}$. Additionally, we feed the same feature vectors $H$ to a color MLP $\boldsymbol{f_{col}}(\cdot)$, to predict the 32-channel color $C$ for each vertex.
The vertex positions to $V_{can}$ and 32-channel color $C$ can be described as:
\begin{align}
 V_{can} &= V_{0} + \delta V_{id} + \lambda_{exp}(V_0) \delta V_{exp} + \lambda_{nf}(V_0) \delta V_{nf}, \\
 C &= \boldsymbol{f_{col}}(H).
\end{align}
$\lambda_{exp}(\cdot)$ and $\lambda_{nf}(\cdot)$ respectively indicate whether the vertices belong to the face area, affected by the facial expression code, or belong to the non-face area, affected by the non-face motion code.


Here, we assume that the vertices closer to the 3D landmarks are more affected by the expression code and less affected by the non-face motion code, while the opposite is true for the vertices far away. Specifically, The 3D landmarks $\boldsymbol{P}_0$ of the canonical model are first estimated through the 3DMM model in the data preprocessing~\ref{subsec:preprocessing} and then optimized later. We calculate the above weight $\lambda_{exp}(\cdot)$ and $\lambda_{nf}(\cdot)$ as follows:
\begin{align*}
\begin{split}
\lambda_{exp}(x)= \left \{
\begin{array}{ll}
    1,                                                               & dist(x, \boldsymbol{P}_0) < t_{1} \\
    \frac{t_{2}-dist(x, \boldsymbol{P}_0)}{t_{2}-t_{1}},               & dist(x, \boldsymbol{P}_0) \in [t_{1},t_{2}]\\
    0,                                                               & dist(x, \boldsymbol{P}_0) > t_{2}
\end{array}
\right.
\end{split}
\end{align*}
with $\lambda_{nf}(x) = 1 - \lambda_{exp}(x)$. And $x \in V_0$ denotes the position of one vertex. $dist(x, \boldsymbol{P}_0)$ denotes the minimum distance from the point $x$ to the 3D landmarks $\boldsymbol{P}_0$. $t_{1}=0.1$ and $t_{2}=0.12$ are predefined hyperparameters when the length of the head is set to approximately $1$.

In practice, we find it difficult and unnecessary to learn expression-related color changes in a generalizable head model, as feeding the expression code to the color MLP may lead to the coupling of appearance and expression. Therefore, we do not consider expression-related color changes during the head model training stage, but model such changes in the downstream reconstruction task (see Section~\ref{subsec:monocular}).

Finally, we utilize the estimated head pose parameters $R$ and $T$ obtained during data preprocessing to transform the mesh from the canonical space to the world space $V = R \cdot V_{can} + T$. After generating the final vertex positions, colors, and faces $\{V, C, F\}$ of the mesh, we render the mesh into a 512-resolution 32-channel feature map $I_{F}$ and a mask $M$ through differentiable rasterization with a given the camera pose. Subsequently, the feature map is interpreted as a 512-resolution fine RGB $I_{fine}$ image through a lightweight convolutional refine network $\boldsymbol{\Psi}(\cdot)$, as shown in Fig.~\ref{fig:method}.

\textbf{3D Gaussian Parametric Head Model.} The Gaussian model also takes an identity code $\boldsymbol{z^{id}}$, a face image $I_{face}$ and a non-face image $I_{nonface}$ as input, producing the positions $X$, color $C$, scale $S$, rotation $Q$ and opacity $A$ of the 3D Gaussians. 
Similar to the guiding geometry model, we initially maintain an overall mean point cloud, with the mean positions $\boldsymbol{X_{0}}$. However, we no longer generate the per-vertex feature $\Gamma$ through $\boldsymbol{f_{mean}}(x)$. Instead, we directly bind it to the Gaussian per-point feature as optimizable variables $\boldsymbol{\Gamma_{0}}$.
This is possible since the number of Gaussian points is fixed at this stage. Then we need to transform the mean point cloud into a neutral-expression point cloud, conditioned by the id code $\boldsymbol{z^{id}}$. To achieve this, we utilize the same injection MLP $\boldsymbol{f_{inj}}(\cdot)$ and identity deformation MLP $\boldsymbol{f_{id}}(\cdot)$ defined in the guiding geometry model, which can generate feature vectors $H = \boldsymbol{f_{inj}}(\boldsymbol{z^{id}}, \boldsymbol{\Gamma_{0}})$ that encode identity information for each point and predict the identity-related displacement $\delta X_{id}$ of each point.
Then, we also need to predict the facial expression conditioned displacement $\delta X_{exp}$, the non-face displacement $\delta X_{nf}$, the resulting positions $X_{can}$ and the 32-channel color $C$ of each point, similar to the approach presented in the guiding geometry model. These can be described as:
\begin{align}
 \delta X_{id} =& \boldsymbol{f_{id}}(H), \\
 \delta X_{exp} =& \boldsymbol{f_{exp}}(H, \boldsymbol{E}_{exp}(I_{face})), \\
 \delta X_{nf} =& \boldsymbol{f_{nf}}(H, \boldsymbol{E}_{nf}(I_{nonface})), \\
 X_{can} =& \boldsymbol{X_{0}} + \delta X_{id} + \lambda_{exp}(\boldsymbol{X_{0}}) \delta X_{exp} + \\ & \lambda_{nf}(\boldsymbol{X_{0}}) \delta X_{nf}, \\
\label{eqn:gaussian_color}
 C =& \boldsymbol{f_{col}}(H).
\end{align}

Unlike the representations of SDF and DMTet, Gaussians have additional attributes that need to be predicted. Here, we introduce a new MLP to predict Gaussian attributes in the canonical space, including the scale $S$, rotation $Q_{can}$, and opacity $A$. In order to ensure the stability of the generated results, we refrain from directly predicting these values. Instead, we predict their offsets $\{\delta S, \delta Q, \delta A\}$ relative to the overall mean values $\{\boldsymbol{S_{0}}, \boldsymbol{Q_{0}}, \boldsymbol{A_{0}}\}$:
\begin{equation}
\label{eqn:gaussian_attributes}
 \{S, Q_{can}, A\} = \{\boldsymbol{S_{0}}, \boldsymbol{Q_{0}}, \boldsymbol{A_{0}}\} + \boldsymbol{f_{att}}(H).
\end{equation}
Also at this stage, we do not consider expression-related Gaussian attributes changes as color changes mentioned above.

Following this, we utilize the estimated head pose parameters $R$ and $T$, obtained during data preprocessing, to transform the canonical space variables $X_{can}$ and $Q_{can}$ into the world space: $X = R \cdot X_{can} + T, \ Q = R \cdot Q_{can}$. For model rendering, we leverage differentiable rendering~\cite{kerbl3Dgaussians} and neural rendering techniques to generate images. The generated 3D Gaussian parameters, which include $\{X, C, S, Q, A\}$, are conditioned by the identity code $\boldsymbol{z^{id}}$, the face image $I_{face}$ and the non-face image $I_{nonface}$. Finally, we input this feature map into the same refine network $\boldsymbol{\Psi}(\cdot)$ of the guiding geometry model to generate a 512-resolution RGB image.

In the 3D Gaussian Parametric Head Model, we leverage the previously trained guiding geometry model to initialize our variables and networks, rather than initiating them randomly and training from scratch. Specifically, we initialize the Gaussian positions $\boldsymbol{X_{0}}$ using the vertex positions of the mean mesh $V_{0}$. Meanwhile, we generate the per-vertex feature $\Gamma$ from $\boldsymbol{f_{mean}}(x)$ at the beginning and bind it to the points as an optimizable variable $\boldsymbol{\Gamma_{0}}$ as described above. Additionally, all identity codes $\boldsymbol{z^{id}}$, 3D landmarks $\boldsymbol{P}_0$ and the networks $\boldsymbol{E}_{exp}(\cdot)$, $\boldsymbol{E}_{nf}(\cdot)$, $\{\boldsymbol{f_{inj}}(\cdot), \boldsymbol{f_{id}}(\cdot), \boldsymbol{f_{exp}}(\cdot), \boldsymbol{f_{nf}}(\cdot), \boldsymbol{f_{col}}(\cdot), \boldsymbol{\Psi}(\cdot)\}$ are directly inherited  from the guiding geometry model. Note that, the attribute MLP $\boldsymbol{f_{att}}(\cdot)$ is a newly introduced network, hence it is initialized randomly. Finally, the overall mean values of the Gaussian attributes $\{\boldsymbol{S_{0}}, \boldsymbol{Q_{0}}, \boldsymbol{A_{0}}\}$ are initialized following the original 3D Gaussian Spatting~\cite{kerbl3Dgaussians}.

\subsection{Loss Functions} 
\label{subsec:loss_functions}
To ensure the accurate convergence of the model, we employ various loss functions as constraints, including the basic photometric loss and silhouette loss, to enforce consistency with ground truth of both the rendered fine images $I_{fine}$ and the rendered masks $M$:
\begin{align}
 \mathcal{L}_{fine} = ||I_{fine} - I_{gt}||_{1}, \\ 
 \mathcal{L}_{sil} = IOU(M, M_{gt}),
\end{align}
with $I_{gt}$ representing the ground truth RGB images, $M_{gt}$ representing the ground truth masks. We further encourage the first three channels of the feature map $I_{coarse}$ to closely match the ground-truth RGB image $I_{gt}$ by introducing an $L_1$ loss:
\begin{equation}
\label{eqn:L_lr}
 \mathcal{L}_{coarse} = ||I_{coarse} - I_{gt}||_{1}.
\end{equation}

The geometric deformation caused by expressions is typically complex and cannot be learned through image supervision alone. Therefore, we provide additional coarse supervision for expression deformation learning using 3D landmarks. Specifically, we define the 3D landmarks $\boldsymbol{P_{0}}$ in the canonical space, and then predict their displacements and transform them to the world space as $\boldsymbol{P}$ just like the transformation of the original vertices $V_{0}$ above. Then, we construct the landmark loss function:
\begin{equation}
\label{eqn:L_lmk}
 \mathcal{L}_{lmk} = ||\boldsymbol{P} - P_{gt}||_{2},
\end{equation}
with $P_{gt}$ denoting the ground truth 3D landmarks, which are estimated by fitting a BFM model to the training data during preprocessing.

Moreover, to guarantee the decoupling of identity deformations and motion deformations learned by the model and minimize redundancy, we introduce the following regularization loss function that aims to minimize the magnitude of motion deformations:
\begin{equation}
\label{eqn:L_reg}
 \mathcal{L}_{reg} = ||\delta V_{exp}||_{2} + ||\delta V_{nf}||_{2}.
\end{equation}

During the training of the \textbf{Guiding Geometry Model}, we also construct a Laplacian smooth term $\mathcal{L}_{lap}$ to penalize surface noise or breaks. Overall, the total loss function is formulated as:
\begin{align}
 \mathcal{L} =& \mathcal{L}_{fine} + \lambda_{sil}\mathcal{L}_{sil} + \lambda_{coarse}\mathcal{L}_{coarse} + \\
 &\lambda_{lmk}\mathcal{L}_{lmk} + 
  \lambda_{reg}\mathcal{L}_{reg} + \lambda_{lap}\mathcal{L}_{lap}
\label{equ:guiding}
\end{align}
with all the $\lambda$ denoting the weights of each term. In practice, we set $\lambda_{sil}=0.1$, $\lambda_{coarse}=0.1$, $\lambda_{lmk}=0.1$, $\lambda_{reg}=0.001$ and $\lambda_{lap}=100$.
During training, we jointly optimize the bolded variables above: \{$\boldsymbol{z^{id}}$, $\boldsymbol{E}_{exp}(\cdot)$, $\boldsymbol{E}_{nf}(\cdot)$, $\boldsymbol{f_{inj}}(\cdot)$, $\boldsymbol{f_{mean}}(\cdot)$, $\boldsymbol{f_{id}}(\cdot)$, $\boldsymbol{f_{exp}}(\cdot)$, $\boldsymbol{f_{nf}}(\cdot)$, $\boldsymbol{f_{col}}(\cdot)$, $\boldsymbol{\Psi}(\cdot)$, $\boldsymbol{P_{0}}$\}. Notably, the defined canonical 3D landmarks $\boldsymbol{P_{0}}$ are initialized by computing the average of the estimated 3D landmarks from the training dataset.

During the training stage of the \textbf{3D Gaussian Parametric Head Model}, we also calculate the perceptual loss~\cite{zhang2018the} to encourage the model to learn more high-frequency details $\mathcal{L}_{vgg} = VGG(I_{fine}, I_{gt})$.
Similar to training the guiding geometry model, we enforce the first three channels of the feature map to be RGB channels as Eqn.~\ref{eqn:L_lr}, introduce landmarks guidance terms as Eqn.~\ref{eqn:L_lmk} and the regular term for the displacement of points as Eqn.~\ref{eqn:L_reg}. Consequently, the overall loss function can be formulated as:
\begin{align}
\label{eqn:gaussian}
 \mathcal{L} =& \mathcal{L}_{fine} + \lambda_{vgg}\mathcal{L}_{vgg} + \lambda_{coarse}\mathcal{L}_{coarse} + \\
 &\lambda_{lmk}\mathcal{L}_{lmk} + \lambda_{reg}\mathcal{L}_{reg}
\end{align}
with the weights $\lambda_{vgg}=0.1$, $\lambda_{coarse}=0.1$, $\lambda_{lmk}=0.1$ and $\lambda_{reg}=0.001$.
In this training stage, we also jointly optimize all the bolded variables and networks mentioned above, including the overall mean positions and attributes of the Gaussians and the 3D landmarks: \{$\boldsymbol{z^{id}}$, $\boldsymbol{E}_{exp}(\cdot)$, $\boldsymbol{E}_{nf}(\cdot)$, $\boldsymbol{f_{inj}}(\cdot)$, $\boldsymbol{f_{id}}(\cdot)$, $\boldsymbol{f_{exp}}(\cdot)$, $\boldsymbol{f_{nf}}(\cdot)$, $\boldsymbol{f_{col}}(\cdot)$, $\boldsymbol{f_{att}}(\cdot)$, $\boldsymbol{\Psi}(\cdot)$, $\boldsymbol{X_{0}}$, $\boldsymbol{\Gamma_{0}}$, $\boldsymbol{S_{0}}$, $\boldsymbol{Q_{0}}$, $\boldsymbol{A_{0}}$, $\boldsymbol{P_{0}}$\}.

\subsection{Training Details}
\label{subsec:training_details}
Before training starts, we first initialize the identity codes following NPHM~\cite{giebenhain2023nphm}. For each different identity, we set a different identity code with a dimension of 512. In addition, we enforce a constraint to ensure that the norm of these codes remains below 1.

Besides, we experimentally found that although it is not necessary, jointly optimizing the head pose during the training process can eliminate some minor errors generated during the BFM calibration process, promoting the consistency of the model as the codes change. Consequently, we optimize all the head poses with a very small learning rate $1 \times 10^{-5}$ throughout the entire training stage of the model.

For the face image in the training process, we randomly select the image of another view of the current frame, or the synthesized image by LivePortrait~\cite{guo2024liveportrait} with the same expression as the input.

Next, we give a brief description of the network structure. The each of the 2 encoders $\boldsymbol{E}_{exp}(\cdot)$, $\boldsymbol{E}_{nf}(\cdot)$ consists of 4 convolution blocks plus an MLP. Each block contains a convolutional layer with stride 1 and a convolutional layer with stride 2. For the refine network $\boldsymbol{\Psi}(\cdot)$, we adopt a very simple U-net~\cite{ronneberger2015unet} structure, with 2 convolutional layer for downsampling and 2 convolutional layers for upsampling. For the MLPs $\boldsymbol{f_{mean}}(\cdot)$, $\boldsymbol{f_{id}}(\cdot)$, $\boldsymbol{f_{exp}}(\cdot)$, $\boldsymbol{f_{nf}}(\cdot)$, $\boldsymbol{f_{col}}(\cdot)$, $\boldsymbol{f_{att}}(\cdot)$, we set the width to 512 with 4 hidden layers. And for the injection MLP $\boldsymbol{f_{inj}}(\cdot)$, we set the width to 512 with 8 hidden layers. The mesh and the Gaussians are both rendered as 512-resolution feature maps and transferred into 512-resolution RGB images through the refine network.

During training the guiding geometry model, we use 256-resolution tetrahedral grid for extracting the mesh via DMTet. For the optimization, we use an Adam~\cite{diederik2015adam} optimizer, and set the learning rate to $1 \times 10^{-4}$ for the identity codes $\boldsymbol{z^{id}}$, $1 \times 10^{-4}$ for all the networks $\boldsymbol{E}_{exp}(\cdot)$, $\boldsymbol{E}_{nf}(\cdot)$, $\boldsymbol{f_{inj}}(\cdot)$, $\boldsymbol{f_{mean}}(\cdot)$, $\boldsymbol{f_{id}}(\cdot)$, $\boldsymbol{f_{exp}}(\cdot)$, $\boldsymbol{f_{nf}}(\cdot)$, $\boldsymbol{f_{col}}(\cdot)$, $\boldsymbol{\Psi}(\cdot)$, and $1 \times 10^{-4}$ for the 3D landmarks $\boldsymbol{P_{0}}$. We use a batch size of 8, with each batch containing 4 images of a specific expression from a given identity. Training the guiding geometry model requires 8 RTX4090 graphics cards and approximately 1 day.

While training the Gaussian model, we also use an Adam optimizer and set the learning rate: $1 \times 10^{-5}$ for the identity codes $\boldsymbol{z^{id}}$, $1 \times 10^{-4}$ for all the networks $\boldsymbol{E}_{exp}(\cdot)$, $\boldsymbol{E}_{nf}(\cdot)$, $\boldsymbol{f_{id}}(\cdot)$, $\boldsymbol{f_{exp}}(\cdot)$, $\boldsymbol{f_{nf}}(\cdot)$, $\boldsymbol{f_{col}}(\cdot)$, $\boldsymbol{f_{att}}(\cdot)$, $\boldsymbol{\Psi}(\cdot)$ and $1 \times 10^{-4}$ for the 3D landmarks $\boldsymbol{P_{0}}$. For the mean Gaussian attributes, we set the learning rates as: $1 \times 10^{-5}$ for the positions $\boldsymbol{X_{0}}$, $1 \times 10^{-5}$ for the per-vertex feature $\boldsymbol{\Gamma_{0}}$, $3 \times 10^{-5}$ for the scale $\boldsymbol{S_{0}}$, $1 \times 10^{-5}$ for the rotation $\boldsymbol{Q_{0}}$ and $1 \times 10^{-4}$ for the opacity $\boldsymbol{A_{0}}$. We use a batch size of 8, with each batch containing a single image. Following the training of the guiding geometry model, we transfer its parameters to the Gaussian model and continue training on 8 RTX4090 graphics cards for 7 days until convergence is achieved.

\begin{figure*}[t]
  \centering
  \includegraphics[width=\linewidth]{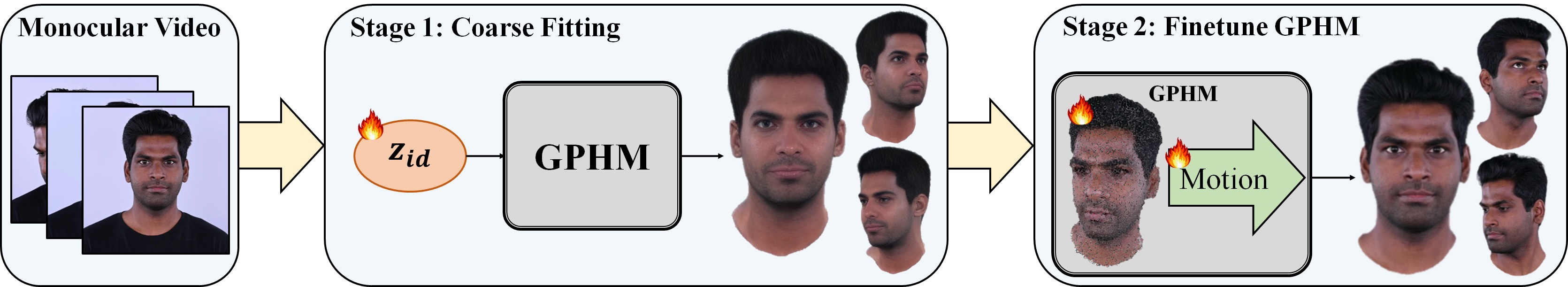}
  \caption{The pipeline of head avatar reconstruction from monocular videos. First, we optimize the identity code $\boldsymbol{z}_{id}$ to coarsely fit the GPHM model to the input video. Then we directly finetune the 3D Gaussian attributes and the motion-related networks in the GPHM for a fine-grained head avatar. The flame chart in the figure marks the parameters that need to be optimized.}
  \label{fig:monocular_recon}
\end{figure*}

\subsection{Head Avatar from a Monocular Video}
\label{subsec:monocular}
Once we have trained the Gaussian Parametric Head Model, one of the main applications is fast reconstruction for 3D head avatars from monocular videos, or even from few-shot or single-image inputs. Previous methods~\cite{xu2023avatarmav, gao2022reconstructing, zielonka2022instant, shao2024splattingavatar, xiang2024flashavatar} trained their model from scratch and used the 3DMM expression coefficients as the driving signal. Therefore, when training the model, sufficient data is required to ensure generalization and reconstruction quality, and the subsequent driving process requires tracking for the 3DMM expression of the source actor. In contrast, we utilize the well-learned generalizable prior model for both appearance and expression to model head avatars. Specifically, we first fit the GPHM to the input video to obtain a rough 3D Gaussian model in a few iterations. Further, we just slightly finetune the 3D Gaussian parameters and the refine network to reconstruct a fine-grained Gaussian model. Our pipeline achieves faster reconstruction and better rendering quality while requiring less training data. In addition, as our framework leverages a pretrained expression encoder for end-to-end expression control, the head avatar can be animated directly by a face image or video, without tracking for 3DMM expression. In the experiment, we also verified that this expression control strategy shows better generalization ability. 

\textbf{GPHM Fitting}
At this stage, we first optimize the identity code to obtain a coarsely fitting GPHM model. Given a $N$ frame monocular frontal portrait video, a set of few-shot images, or even a single image input $\{{I_n}\}, n \in \{1,...,N\}$, we first remove the background of the images, then detect 2D landmarks for extracting face images $\{{I_{face}^n}\}, n \in \{1,...,N\}$, non-face images $\{{I_{nonface}^n}\}, n \in \{1,...,N\}$ and estimating the head pose $\{{R_n, T_n}\}, n \in \{1,...,N\}$ in the same way as the training data preprocessing described in Section~\ref{subsec:preprocessing}. Then, we randomly initialize a global identity code $\boldsymbol{z^{id}}$ and fit our GPHM to the input video by optimizing the identity code. Specifically, in each iteration, we sample one frame $n$ and input the identity code $\boldsymbol{z^{id}}$, the face image ${I_{face}^n}$ and the non-face image ${I_{nonface}^n}$ to the GPHM to generate 3D Gaussians $\{X_n, C_n, S_n, Q_n, A_n\}$ as described above. We then render the feature map $I_{F}$ through rasterization with the first three channels as the coarse RGB image $I_{coarse}$. Finally the feature map $I_{F}$ is transferred to fine image $I_{fine}$ through the refine network $\boldsymbol{\Psi}$. For the loss function, we use only photometric loss $\mathcal{L}_{coarse}$ and $\mathcal{L}_{fine}$ defined in Eqn.~\ref{eqn:gaussian}. We totally optimize the identity code $\boldsymbol{z^{id}}$ for 100 iterations with learning rate $1 \times 10^{-3}$.

\textbf{Finetune 3D Gaussians}
In this stage, we further optimize the Gaussian attributes to reconstruct a high-fidelity identity-specific head avatar. 
First, we calculate the Gaussians positions $\boldsymbol{X_{id}} = \boldsymbol{X_{0}} + \delta X_{id}$, the color and other Gaussian attributes $\boldsymbol{C_{id}}$, $\boldsymbol{S_{id}}$, $\boldsymbol{Q_{id}}$, $\boldsymbol{A_{id}}$ as Eqn.~\ref{eqn:gaussian_color} and Eqn.~\ref{eqn:gaussian_attributes} using the identity code obtained in the GPHM fitting stage and set them as optimizable variables. As the computations are completed, we no longer need the networks $\boldsymbol{f_{inj}}(\cdot)$, $\boldsymbol{f_{id}}(\cdot)$, $\boldsymbol{f_{col}}(\cdot)$, $\boldsymbol{f_{att}}(\cdot)$ and the identity code. So they are subsequently discarded for saving computational overhead.
Then, we consider modeling some identity-specific dynamic details deriving from the color and Gaussian attributes varying with the expression. Referring to the approach in Gaussian Head Avatar~\cite{xu2024gaussian}, we introduce an additional tiny MLP $\boldsymbol{f_{dyn}}(\cdot)$, which takes the facial expression code $z^{exp}$ and the per-Gaussian feature $\gamma$ as input and predicts the offset of the variables $\{\delta C, \delta S, \delta Q, \delta A\}$. 
Finally, the optimization process is the same as the GPHM Fitting stage. In each iteration, one frame $n$ is sampled, the face image ${I_{face}^n}$ and the non-face image ${I_{nonface}^n}$ are input to the model and render the feature map $I_{F}$ which is transferred to the fine image $I_{fine}$ later. Finally, we construct the same loss function as Eqn.~\ref{eqn:gaussian}.
Typically, during finetuning, we optimize the identity-specific Gaussians $\boldsymbol{X_{id}}$, $\boldsymbol{C_{id}}$, $\boldsymbol{S_{id}}$, $\boldsymbol{Q_{id}}$, $\boldsymbol{A_{id}}$ with learning rate $1 \times 10^{-3}$ and the networks $\boldsymbol{f_{exp}}(\cdot)$, $\boldsymbol{\Psi}(\cdot)$, $\boldsymbol{f_{dyn}}(\cdot)$ with learning rate $3 \times 10^{-4}$ for 2000 iterations. But in the case of few-shot input, the optimization of the facial expression MLP $\boldsymbol{f_{exp}}(\cdot)$ and refine network $\boldsymbol{\Psi}(\cdot)$ is optionally turned off to prevent overfitting and reduce the number of iterations appropriately according to the number of input images.

\textbf{Reenactment}
Once the identity-specific head avatar is reconstructed, we can use another portrait video for reenactment. Given a frame of the driving video, we first estimate the head pose and extract face image and non-face image as training data preprocessing~\ref{subsec:preprocessing}. Then, we feed the face image to the facial expression encoder to generate the facial expression code for expression controlling. Similarly, we also feed the non-face image to the non-face motion encoder to generate non-face motion code to control the neck and shoulder motion when turning the head. Note, optionally the control of the neck and below could be ignored  by fixing the non-face motion code as a sample in the training set. Finally, we input two motion codes to the finetuned model to generate 3D Gaussians, which is rendered into images.

\section{Experiments}
\label{sec::experiments}

\subsection{Datasets}

\textbf{NeRSemble} dataset contains over 260 different identities, and collects 72fps multi-view videos from 16 synchronized cameras for each identity. The combined video frames for each identity range from approximately 6000 to 11000 frames. For each identity video, we selected about 800 frames from all 16 views as training data.

\textbf{NPHM} dataset contains 5200 3D human head scans. These scans come from 255 different identities, each with about 20 different expressions. Since our method utilizes 2D images as training supervision, we render each scan from 80 different views to generate synthetic image data and record the camera parameters and the masks.

\textbf{FaceVerse} dataset is an East Asian human head scan dataset. It contains 2310 scans from 110 different identities, and each identity contains 21 expressions. We selected 1620 scan data of 80 identities for training. Similarly, for each scan, we render multi-view synthetic image data from 80 different views and record the camera parameters and the masks.

\textbf{VFHQ} dataset is a large-scale high-quality monocular video dataset, containing interviews and speeches of various people, with a total of 15,000+ videos, each with hundreds of frames. We removed some clips with poor quality and poor background segmentation, and selected 6,000 videos for training our model.

\subsection{Evaluation for Gaussian Parametric Model.}

\begin{figure}[t]
  \centering
  \includegraphics[width=1.0\linewidth]{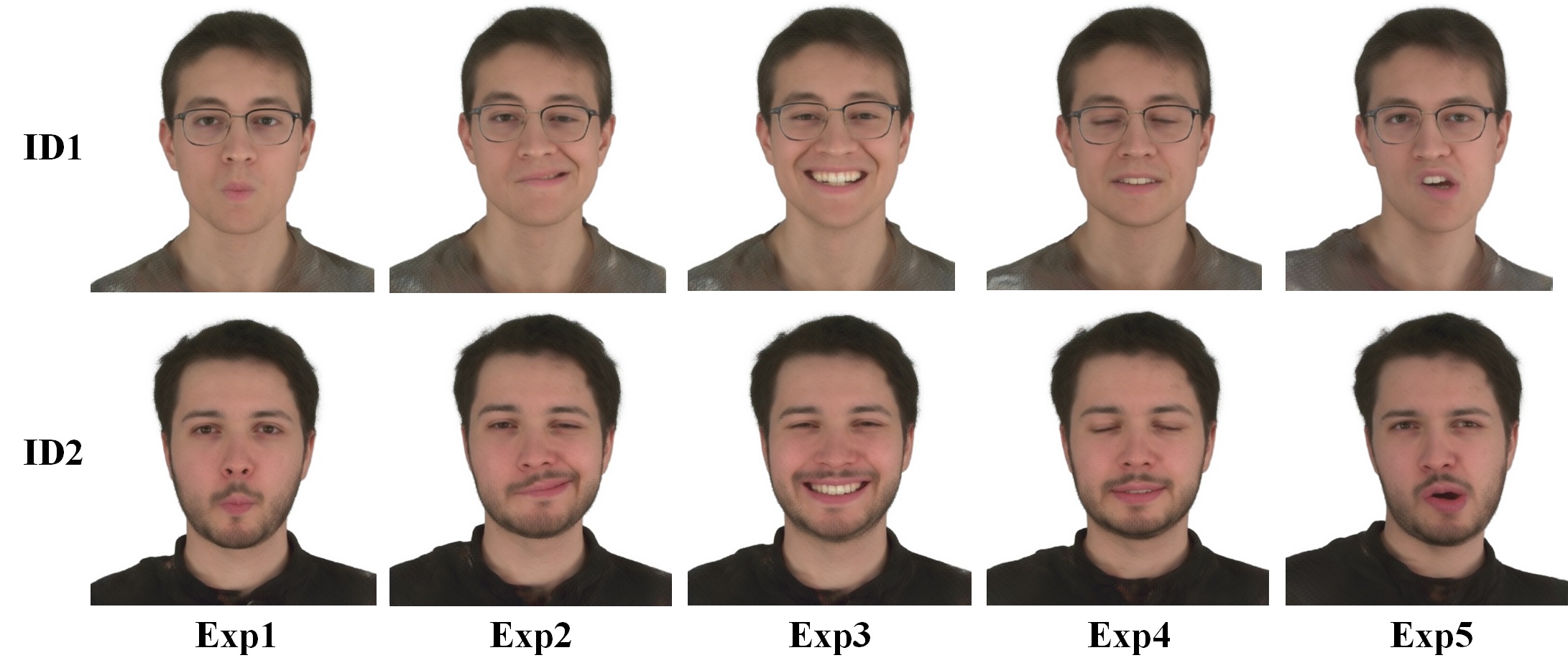}
  \caption{We generate the head models with randomly sampled identity codes and expression codes as conditions. Each row corresponds to the same identity code, and each column corresponds to the same expression code.}
  \label{fig:samples}
\end{figure}

\textbf{Disentanglement.}
We tested the performance of the 3D Gaussian Parametric Model under the control of different identity codes and different expression codes. We randomly sampled 2 identity codes and 5 expression codes to generate 10 head models. Each horizontal row corresponds to the same identity code, and each column corresponds to the same expression code, as shown in Fig.~\ref{fig:samples}. It can be observed that our model performs well in identity consistency and expression consistency, and the two components are fully disentangled.

\begin{figure}[t]
  \centering
  \includegraphics[width=1.0\linewidth]{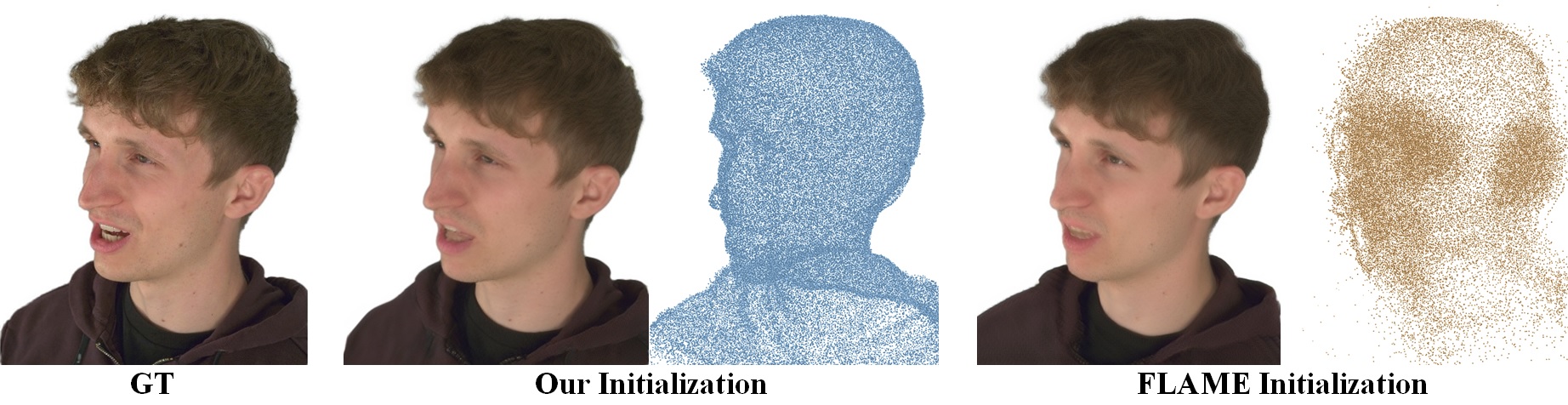}
  \caption{We compare our initialization strategy with using the vertices of FLAME model. The left side shows the rendered image, and the right side shows the positions of the Gaussian points.}
  \label{fig:ablation_initializaiton}
\end{figure}

\textbf{Ablation on Initialization.}
To evaluate the effectiveness of our initialization strategy with guiding geometry model outlined in Section~\ref{sec:method}, we compare it against a FLAME-based initialization strategy. To use FLAME model for the initialization, we first fit a FLAME model to overall mean 3D landmarks which are estimated during data preprocessing. Then, we sample 100,000 points near the surface of the FLAME mesh as an initialization of the mean Gaussian positions $\boldsymbol{X_{0}}$. For the per-vertex features bound to each point $\boldsymbol{\Gamma}$, we just set them to zero. And for all the networks $\{\boldsymbol{f_{inj}}(\cdot), \boldsymbol{f_{id}}(\cdot), \boldsymbol{f_{exp}}(\cdot), \boldsymbol{f_{col}}(\cdot), \boldsymbol{\Psi}(\cdot)\}$ and $\boldsymbol{f_{att}}(\cdot)$ are randomly initialized as there is no available prior. The initialization process for the Gaussian attributes $\{\boldsymbol{S_{0}}, \boldsymbol{Q_{0}}, \boldsymbol{A_{0}}\}$ remains the same as in our strategy.

We show the visualization results in Fig.~\ref{fig:ablation_initializaiton}, with the Gaussian model rendering image on the left and the Gaussian positions displayed as point clouds on the right. Our initialization strategy using the guiding geometry model can ensure that all the Gaussian points fall evenly on the actual surface of the model, thereby ensuring reconstruction quality. When using the FLAME model for the initialization, a large number of points wander inside or outside the actual surface of the model, causing noise or redundancy and leading the model to lose some high-frequency information and making it difficult to fully converge. We also perform a quantitative evaluation of different initialization strategies on the rendered images, as shown in Table~\ref{tab:ablation}, which shows that our method leads to better rendering results.

\begin{table}
\centering
\setlength{\tabcolsep}{11pt}
\begin{tabular}{c|c|c|c}
\hline
Method                   & PSNR $\uparrow$    & SSIM $\uparrow$    & LPIPS $\downarrow$  \\
\hline
\hline
FLAME Initialization     & 25.7              & 0.82                & 0.109                \\
Our Initialization       & \textbf{28.0}     & \textbf{0.84}       & \textbf{0.085}        \\
\hline
\end{tabular} 
\caption{Quantitative evaluation results of our initialization strategy and naive FLAME initialization strategy.}
\label{tab:ablation}
\end{table}

\begin{figure}[t]
  \centering
  \includegraphics[width=1.0\linewidth]{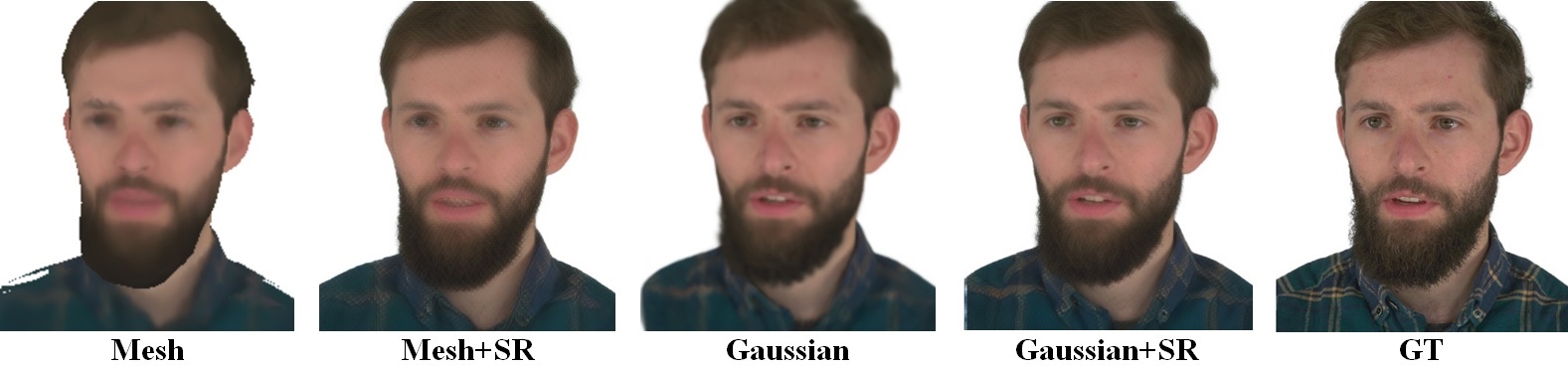}
  \caption{The comparison of the different representations with super-resolution.}
  \label{fig:add_ablation}
\end{figure}

\textbf{Ablation on Representation and Super Resolution.}
We conduct the ablation study for the guiding mesh model, the Gaussian model, and the super-resolution network (abbreviated as SR) as shown in Fig.~\ref{fig:add_ablation}. 
The corresponding PSNR metrics are: Mesh (15.7), Mesh+SR (17.3), Gaussian (27.0), Gaussian+SR (29.3). Compared to mesh, utilizing 3D Gaussian as the representation brings significant improvements (+12), while the super-resolution module adds some details, generating more realistic results.

\subsection{Applications: Head Avatar from a Monocular Video.}
\begin{figure*}[t]
  \centering
  \includegraphics[width=1.0\linewidth]{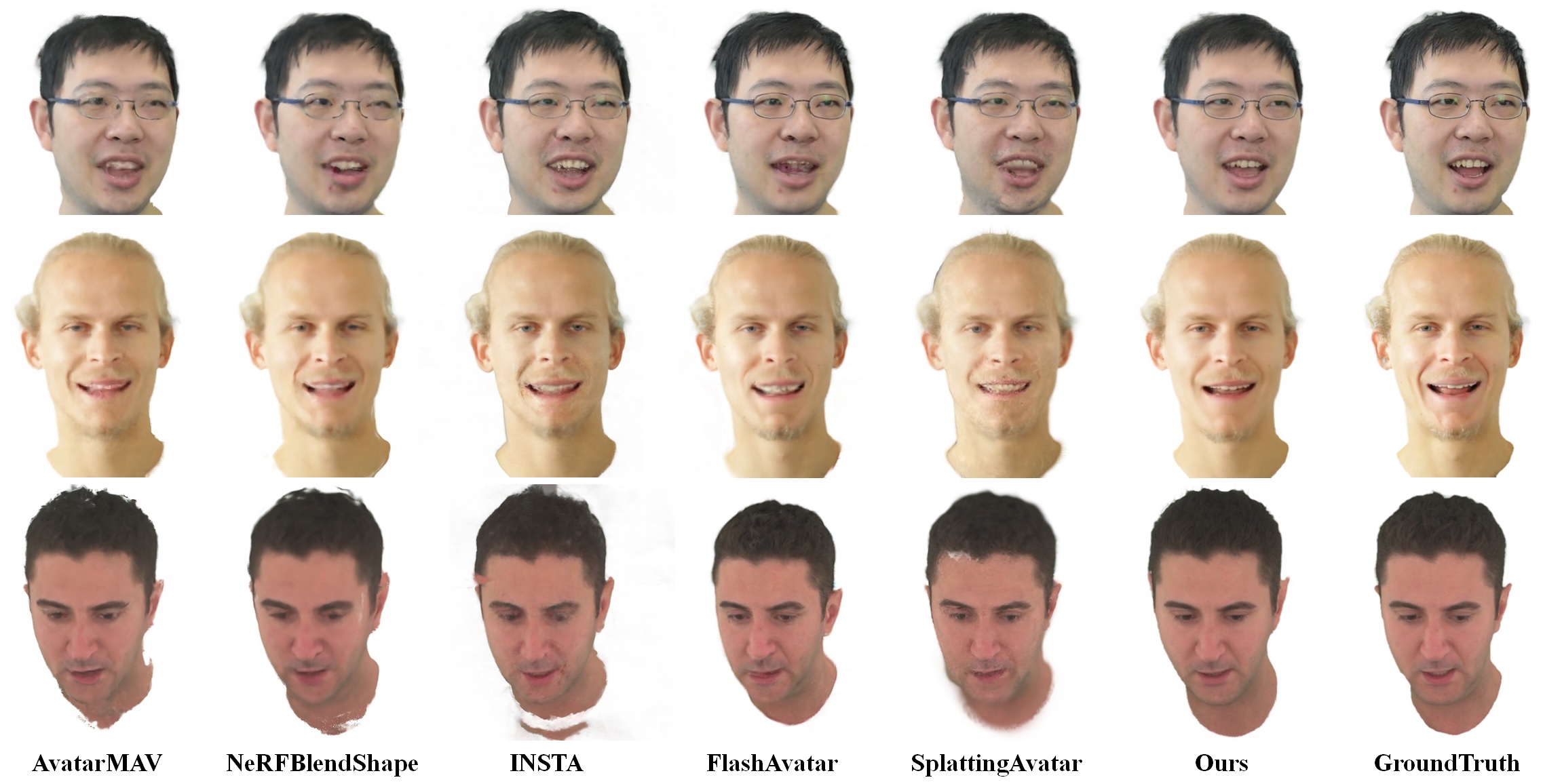}
  \caption{Qualitative comparison of our method and 5 other state-of-the-art methods on self reenactment task. From left to right: AvatarMAV~\cite{xu2023avatarmav}, NeRFBlendShape~\cite{gao2022reconstructing}, INSTA~\cite{zielonka2022instant}, FlashAvatar~\cite{xiang2024flashavatar}, SplattingAvatar~\cite{shao2024splattingavatar} and Ours.}
  \label{fig:self_reenactment}
\end{figure*}

\textbf{Self Reenactment.}
We conduct qualitative and quantitative rendering quality comparisons between our method and five other SOTA monocular head avatar reconstruction methods on self reenactment task. Among them, AvatarMAV~\cite{xu2023avatarmav}, NeRFBlendShape~\cite{gao2022reconstructing} and INSTA~\cite{zielonka2022instant} utilize Voxel-based representation for NeRF head avatar training acceleration. FlashAvatar~\cite{xiang2024flashavatar} and SplattingAvatar~\cite{shao2024splattingavatar} introduce 3D Gaussians and bind them to a FLAME template to model the head avatars. In the experiment, we input 1-minute videos as training data and use additional 20-second videos as evaluation data. And we train their models according to the time claimed by each method. The training time of our method is 5 minutes. The qualitative results are shown in the Fig.~\ref{fig:self_reenactment}. Our method is significantly better than other methods in terms of rendering quality, expression transfer accuracy and robustness. Table.~\ref{tab:self_reenactment} shows the quantitative evaluation results. We evaluate these methods on three metrics: Peak Signal-to-Noise Ratio (PSNR), Structure Similarity Index (SSIM) and Learned Perceptual Image Patch Similarity (LPIPS). Our method achieves the best results on all the metrics and significantly outperforms other SOTA methods on the LPIPS metric.

\begin{table}
\centering
\setlength{\tabcolsep}{11pt}
\begin{tabular}{c|c|c|c}
\hline
Method                           & PSNR $\uparrow$    & SSIM $\uparrow$     & LPIPS $\downarrow$ \\
\hline
\hline
AvatarMAV                        & 27.7               & 0.92                & 0.081              \\
NeRFBlendShape                   & 27.9               & 0.92                & 0.085              \\
INSTA                            & 27.5               & 0.92                & 0.076              \\
FlashAvatar                      & 28.8               & 0.93                & 0.051              \\
SplattingAvatar                  & 28.4               & 0.93                & 0.065              \\
Ours (wo FT motion)              & 28.1               & 0.93                & 0.046              \\
Ours (wo $\boldsymbol{f_{dyn}}$) & 28.2               & 0.93                & 0.044              \\
Ours                             & \textbf{28.9}      & \textbf{0.94}       & \textbf{0.041}     \\
\hline
\end{tabular} 
\caption{Quantitative evaluation results on the task of self reenactment. We compare our method with other 5 SOTA methods: AvatarMAV~\cite{xu2023avatarmav}, NeRFBlendShape~\cite{gao2022reconstructing}, INSTA~\cite{zielonka2022instant}, FlashAvatar~\cite{xiang2024flashavatar} and SplattingAvatar~\cite{shao2024splattingavatar}. And we also include two ablation baselines: Ours (wo $\boldsymbol{f_{dyn}}$) in which we remove the dynamic generator $\boldsymbol{f_{dyn}}$, and Ours (wo FT motion) in which we only optimize the 3D Gaussians but the motion related networks.}
\label{tab:self_reenactment}
\end{table}

\begin{figure*}[t]
  \centering
  \includegraphics[width=1.0\linewidth]{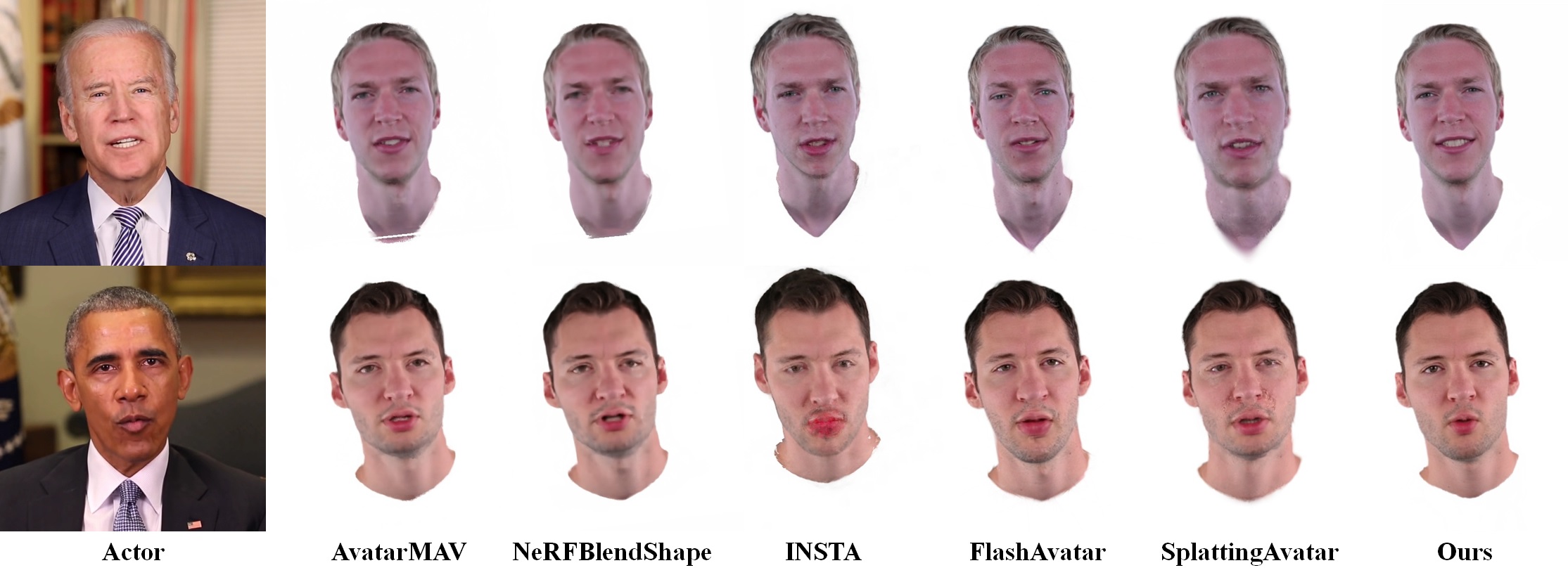}
  \caption{Qualitative comparison of our method and 5 other state-of-the-art methods on cross-identity reenactment task. From left to right: AvatarMAV~\cite{xu2023avatarmav}, NeRFBlendShape~\cite{gao2022reconstructing}, INSTA~\cite{zielonka2022instant}, FlashAvatar~\cite{xiang2024flashavatar}, SplattingAvatar~\cite{shao2024splattingavatar} and Ours.}
  \label{fig:cross_reenactment}
\end{figure*}

\textbf{Cross-identity Reenactment.}
We also compare our method with these state-of-the-art methods on the cross-identity reenactment task. The qualitative results are shown in Fig.~\ref{fig:cross_reenactment}. Other methods suffer from shape and expression coupling because they use a 3DMM model to control their avatars or use the 3DMM expression coefficient as a condition. As a result, the quality of the results is affected by the different shapes when applying cross-identity reenactment. Our method directly inputs the image into a well-decoupled expression encoder which is trained on large-scale datasets to extract latent expressions. Therefore, our method achieves better performance on cross-identity reenactment tasks.

\begin{figure*}[t]
  \centering
  \includegraphics[width=1.0\linewidth]{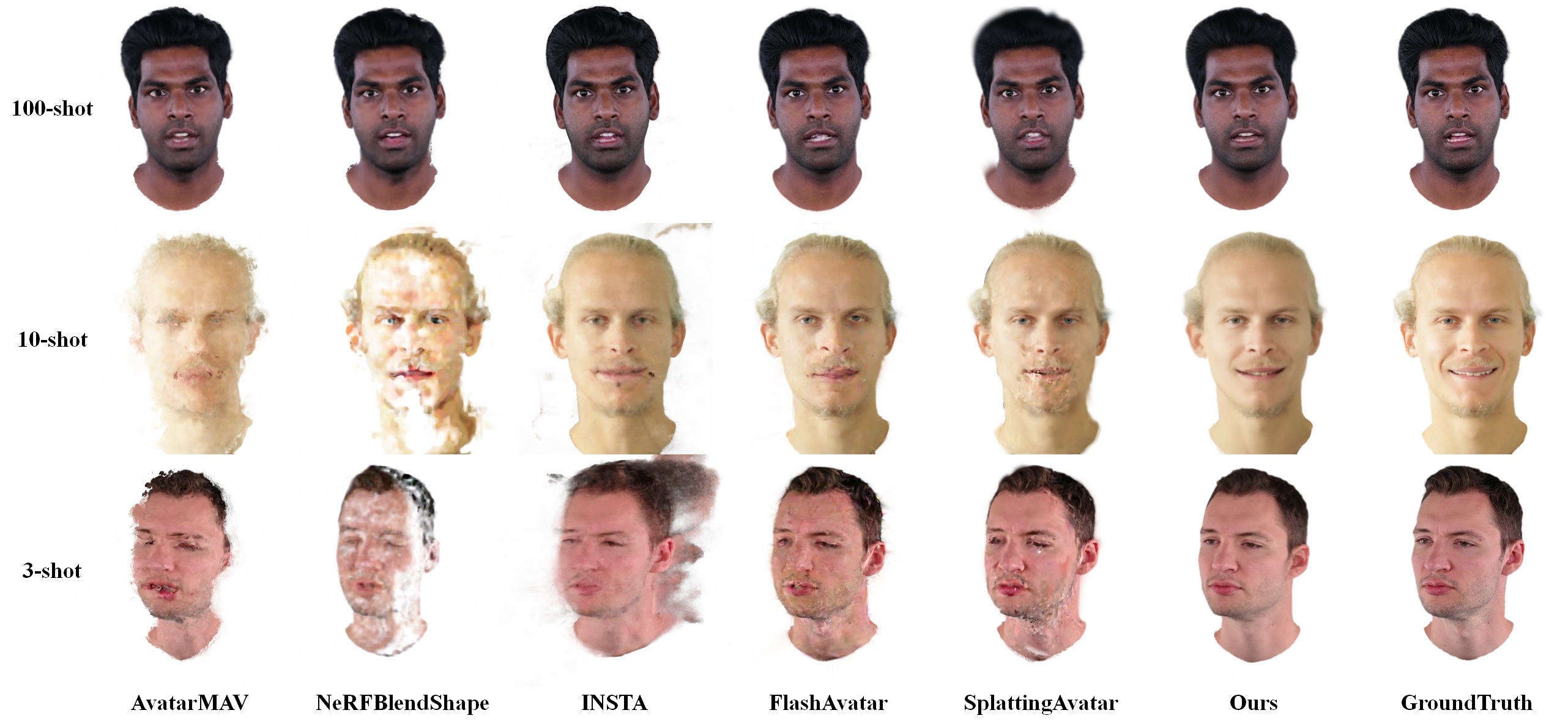}
  \caption{We qualitatively compare our method with 5 other state-of-the-art methods on self reenactment tasks in the 100-shot, 10-shot, and 3-shot cases from top to bottom.  The methods are AvatarMAV~\cite{xu2023avatarmav}, NeRFBlendShape~\cite{gao2022reconstructing}, INSTA~\cite{zielonka2022instant}, FlashAvatar~\cite{xiang2024flashavatar}, SplattingAvatar~\cite{shao2024splattingavatar} and Ours from left to right.}
  \label{fig:fewshot}
\end{figure*}

\begin{figure}[t]
  \centering
  \includegraphics[width=1.0\linewidth]{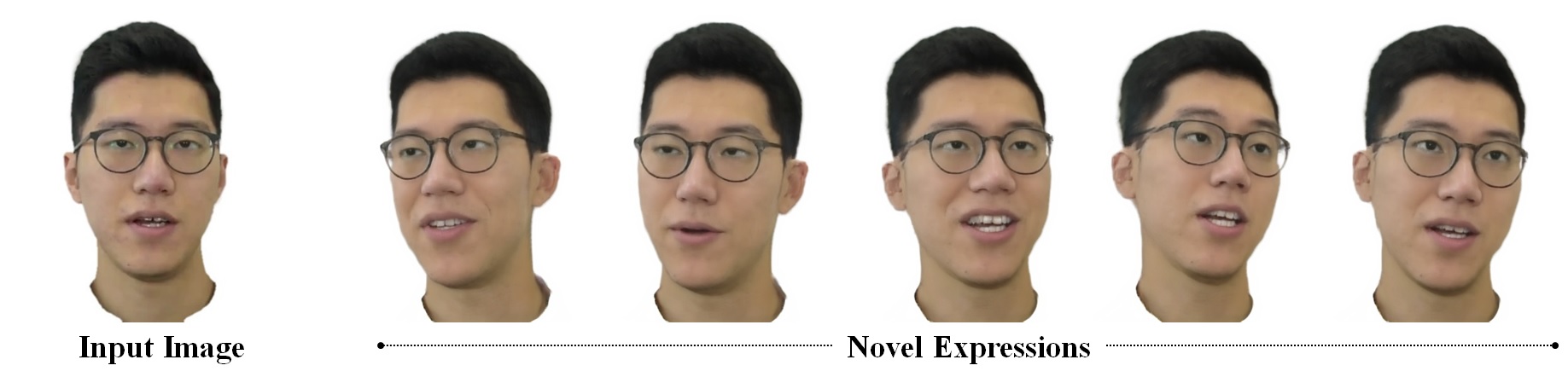}
  \caption{The head avatar reconstruction result of our method when only one single image is used as input.}
  \label{fig:1shot}
\end{figure}

\textbf{Ablation on Few-shot Input.}
Next, we conduct experiments under the setting of few-shot input. We limit the number of input images to 100, 10, and 3 frames, and compare our method with the state-of-the-art methods mentioned above. The qualitative results are shown in Fig.~\ref{fig:fewshot}. While other methods suffer from blurring, artifacts and significant quality degradation as the number of input images decreases, our method can achieve robust and high-quality avatar reconstruction. Even when inputting only one single image, our method can still guarantee robust and high-quality results as shown in Fig.~\ref{fig:1shot}.

\textbf{Ablation on Utilizing Synthesized Expression Condition Images.}

\begin{figure}[t]
  \centering
  \includegraphics[width=1.0\linewidth]{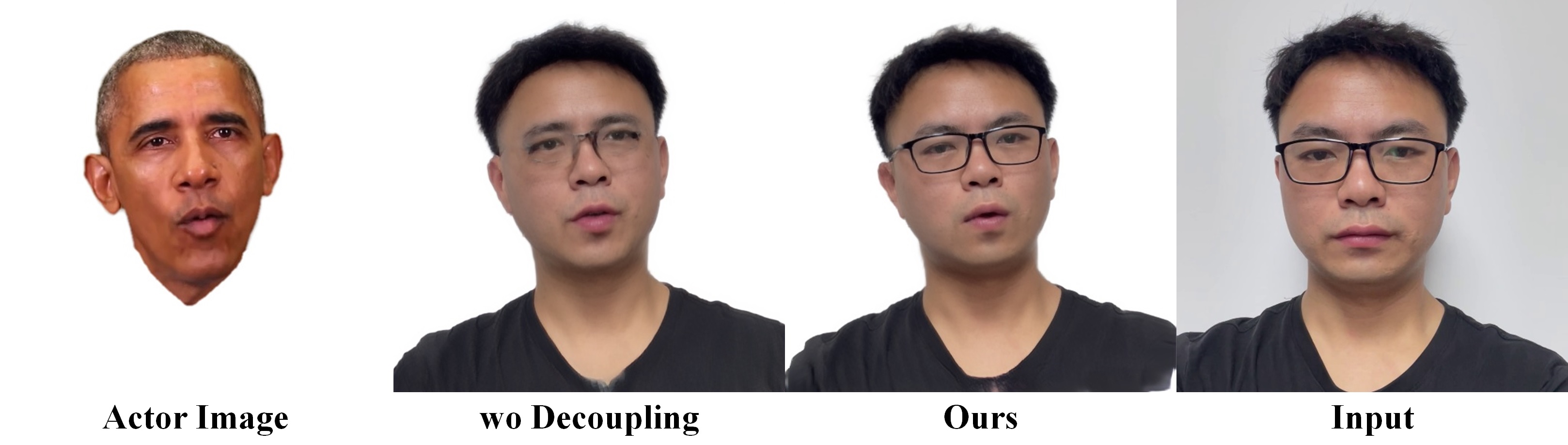}
  \caption{We use synthesized images via LivePortrait to train our expression encoder for expression and appearance decoupling.}
  \label{fig:decoupling}
\end{figure}

As explained in Sec.~\ref{subsec:preprocessing} and Sec.~\ref{subsec:training_details}, for the images input to the facial expression encoder, we utilize LivePortrait~\cite{guo2024liveportrait} to generate additional expression condition images. This strategy forces the encoder to learn only expression information from the input images, thereby achieving expression and appearance decoupling. We also construct an ablation baseline, in which we train the GPHM model using only the groundtruth images as the expression condition and the results are shown in Fig.~\ref{fig:decoupling}. Without utilizing the additional expression condition images, the expression encoder leaks appearance-related information to the motion networks, causing the reconstructed head avatar to still be heavily affected by the actor's appearance during cross-identity reenactment. 

\begin{figure*}[t]
  \centering
  \includegraphics[width=1.0\linewidth]{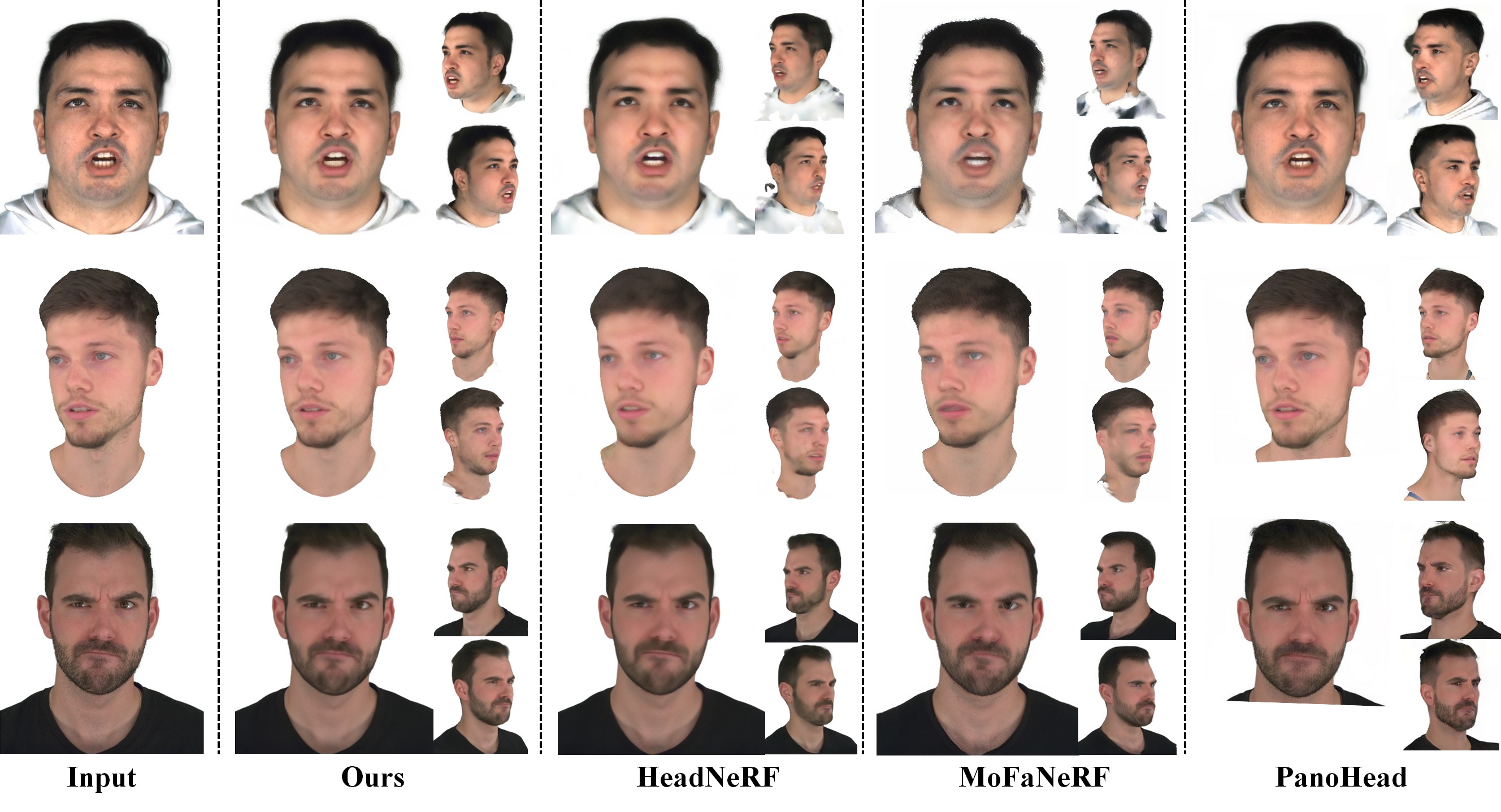}
  \caption{We compare our method with other SOTA methods on the  task of single image fitting. The far left is the input image, and to the right are Our method, HeadNeRF~\cite{hong2022headnerf}, MoFaNeRF~\cite{zhuang2022mofanerf} and PanoHead~\cite{an2023panohead}. Our model significantly outperforms other methods in reconstruction quality and 3D consistency.}
  \label{fig:recon}
\end{figure*}

\subsection{Applications: Image Fitting.}
In this section, we demonstrate the capability of our 3D Gaussian Parametric Model for single-image fitting using the fitting strategy detailed in Section~\ref{subsec:monocular}. We compare our model with similar works: HeadNeRF~\cite{hong2022headnerf}, MoFaNeRF~\cite{zhuang2022mofanerf}, and PanoHead~\cite{an2023panohead}. In addition to evaluating the above methods on our evaluation dataset, we also conduct comparisons using cases from MEAD~\cite{kaisiyuan2020mead} dataset (the first two rows). 
The qualitative results are presented in Fig.~\ref{fig:recon}. Our model exhibits reconstruction accuracy while maintaining excellent 3D consistency and identity preservation. HeadNeRF's fitting results often suffer from missing hair, and they remove the body and neck. MoFaNeRF, trained solely on the FaceScape dataset where all subjects wear hats, struggles to fit hair. As a GAN-based model, PanoHead can achieve highly accurate reproductions from the input view. However, due to overfitting, the results from side views reveal poor 3D consistency and identity preservation. 

In addition to qualitative evaluations, we also conducted quantitative evaluations on 60 images using three metrics: Peak Signal-to-Noise Ratio (PSNR), Structural Similarity Index (SSIM), and Face Distance (FD). Here, we provide a brief explanation of the Face Distance (FD). To compute the FD metric, we utilized a face recognition tool~\footnote{https://github.com/ageitgey/face\_recognition} to encode two images containing faces into 128-dimensional vectors. Subsequently, we calculated the distance between these two vectors to reflect the similarity of the two faces. In our experiments, FD serves as an indicator of identity consistency. The results are shown in Table~\ref{tab:image_based}. Our model demonstrates optimal performance in both fitting accuracy and identity consistency.

\begin{table}
\centering
\setlength{\tabcolsep}{11pt}
\begin{tabular}{c|c|c|c}
\hline
Method             & PSNR $\uparrow$     & SSIM $\uparrow$     & FD $\downarrow$   \\
\hline
\hline
HeadNeRF           & 28.9                & 0.84                & 0.37              \\
MoFaNeRF           & 28.6               & 0.82                & 0.37              \\
PanoHead           & 29.1               & \textbf{0.86}       & 0.41              \\
Ours               & \textbf{30.3}       & \textbf{0.86}       & \textbf{0.35}     \\
\hline
\end{tabular} 
\caption{Quantitative evaluation results on the task of single image fitting. We compare our method with other 3 SOTA methods: HeadNeRF~\cite{hong2022headnerf}, MoFaNeRF~\cite{zhuang2022mofanerf}, PanoHead~\cite{an2023panohead}.}
\label{tab:image_based}
\end{table}

\begin{figure*}[t]
  \centering
  \includegraphics[width=1.0\linewidth]{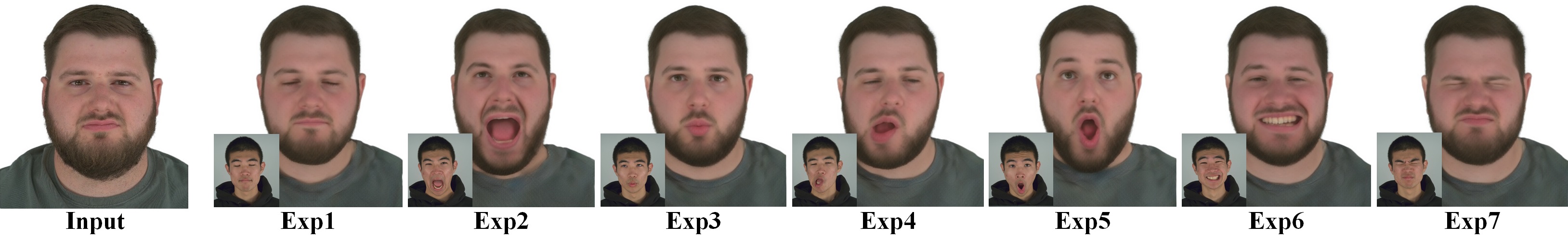}
  \caption{We perform expression editing on the head model reconstructed from the input image. Our model is able to handle very exaggerated expressions with superior identity consistency.}
  \label{fig:reenactment}
\end{figure*}

Our 3D Gaussian Parametric Head Model possesses the capability for expression editing. Upon completing the fitting process on a portrait image, we can animate the model by applying different expression codes. An example is illustrated in Figure~\ref{fig:reenactment}. Our model can generate images depicting the corresponding expressions of the input subject based on a reference expression (as seen in the lower left corner of each image in the figure). It performs admirably even with exaggerated expressions, producing natural and realistic results.

\section{Discussion}

\noindent\textbf{Ethical Considerations.} Our technique can generate artificial portrait videos, posing a significant risk of spreading misinformation, shaping public opinions, and undermining trust in media outlets. These consequences could have profound negative effects on society. Therefore, it is crucial to explore methods that effectively differentiate between genuine and manipulated content.

\noindent\textbf{Limitation.} Our 3D Gaussian Parametric Head Model takes a step forward in the characterization of parametric head models. However, due to the limited amount of training data, the generalization ability of the model is still insufficient. In some cases where the illumination is significantly different from the training set, the reconstruction results are not good.

\noindent\textbf{Conclusion.} In this paper, we propose the 3D Gaussian Parametric Head Model, a novel framework for parametric head model. This model leverages the power of 3D Gaussians, enabling realistic rendering quality and real-time speed. Our well-designed training strategy ensures stable convergence while enabling the model to learn appearance details and expressions. Besides, our model allows for creating detailed, high-quality face avatars from a single input image, and also enables editing for expressions and identity. We believe our model represents a significant advancement in the field of parametric head model. 

\noindent\textbf{Acknowledgment}
The work is supported by the National Natural Science Foundation of China (NSFC) under Grant Number 62125107, 62402274 and the Postdoctoral Fellowship Program of China Postdoctoral Science Foundation under Grant Number GZC20231304.

\bibliographystyle{IEEEtran}
\bibliography{main}

\newpage


\begin{IEEEbiography}[{\includegraphics[width=1in,height=1.25in,clip,keepaspectratio]{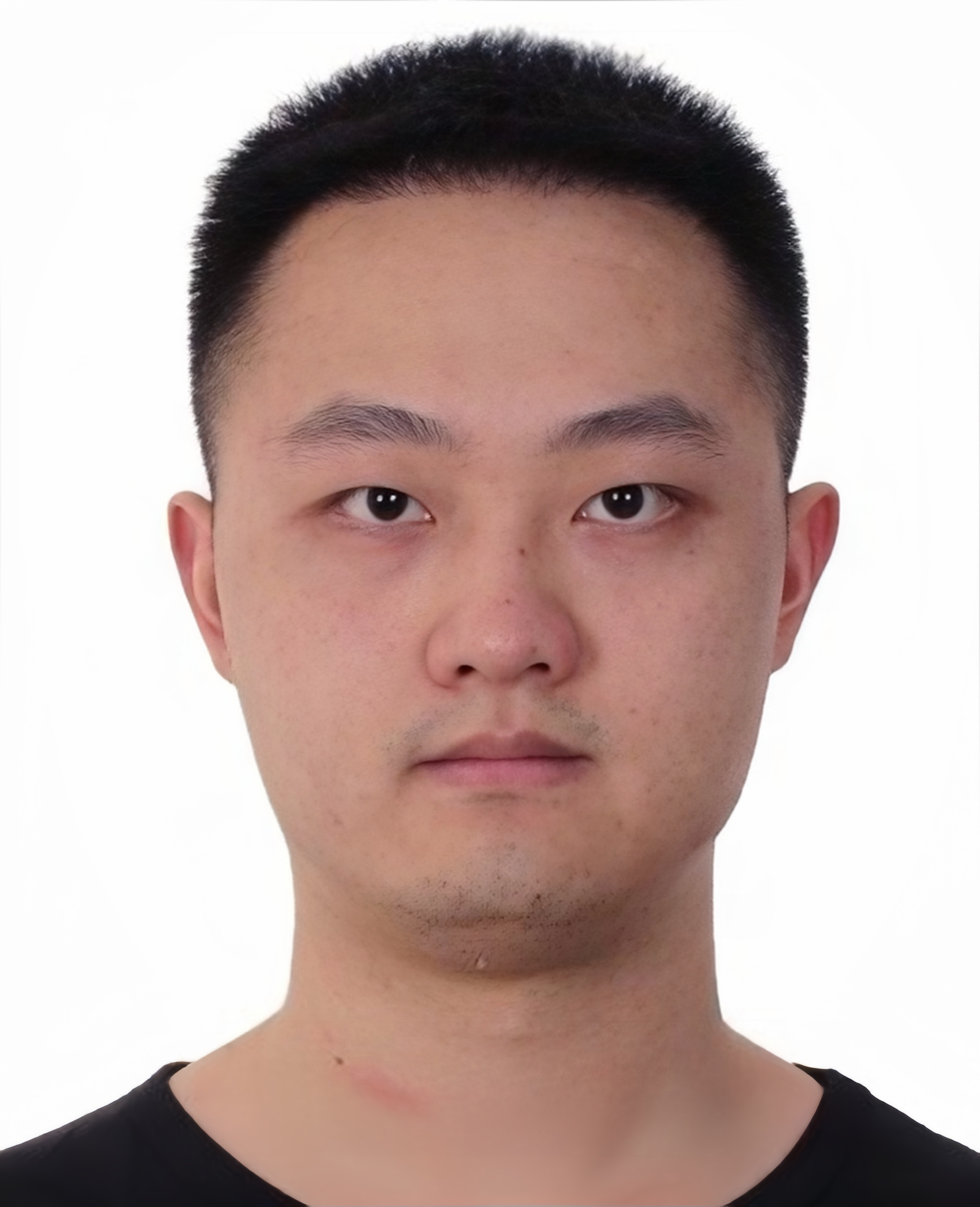}}]{Yuelang Xu}
is currently a 5th year Ph.D. student at Department of Automation, Tsinghua University, advised by Prof. Yebin Liu. His research interests are 3D Computer Vision and Computer Graphics, including 3D head avatar reconstruction and animation, NeRF, 3D Gaussian Splatting.
\end{IEEEbiography}

\begin{IEEEbiography}[{\includegraphics[width=1in,height=1.25in,clip,keepaspectratio]{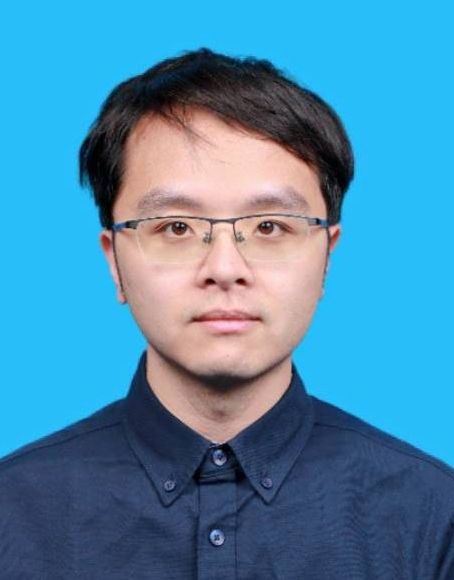}}]{Zhaoqi Su}
is currently a postdoctoral researcher in 3D Vision and Computational Photography Lab, Department of Automation, Tsinghua University. He received the B.S. degree in the Department of Physics, Tsinghua University in 2017, and the Ph.D. degree in the Department of Automation, Tsinghua University in 2023. His current research interests include computer vision and computer graphics, mainly focusing on 3D garment digitization and 3D digital human modeling.
\end{IEEEbiography}

\begin{IEEEbiography}[{\includegraphics[width=1in,height=1.25in,clip,keepaspectratio]{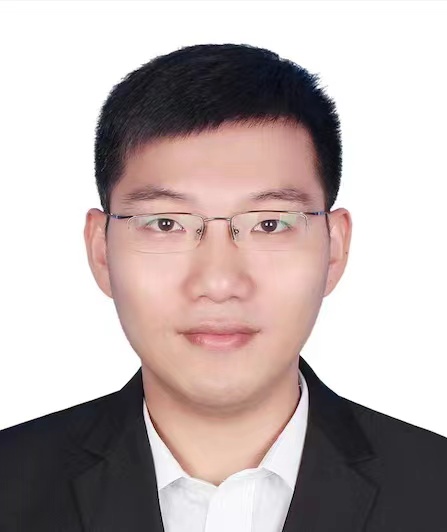}}]{Qingyao Wu} (Member, IEEE) received the B.S. degree in software engineering from the South China University of Technology, China, in 2007, and the Ph.D. degree in computer science from the Harbin Institute of Technology, China, in 2013. He is currently a Professor with the School of Software Engineering, South China University of Technology. His research interests include computer vision and embodied AI.
\end{IEEEbiography}

\begin{IEEEbiography}[{\includegraphics[width=1in,height=1.25in,clip,keepaspectratio]{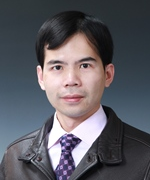}}]{Yebin Liu}
received the B.E. degree from the Beijing University of Posts and Telecommunications, China, in 2002 and the Ph.D. degree from the Automation
Department, Tsinghua University, Beijing, China, in 2009. He is currently a full professor with Tsinghua University. He was a research fellow in the Computer Graphics Group of the Max Planck Institute for Informatik, Germany, in 2010. His research areas include computer vision, computer graphics, and
computational photography.
\end{IEEEbiography}

\vfill

\end{document}